\documentclass[Afour,sageh,times]{sagej}

\usepackage{moreverb,url}
\usepackage{enumitem}
\usepackage{multirow} %multirow for format of table
\usepackage{tabularx}
\usepackage{makecell}
\usepackage{pifont}
\usepackage{gensymb}
\usepackage{amsmath}
\usepackage{balance}
\usepackage[colorlinks,bookmarksopen,bookmarksnumbered,citecolor=red,urlcolor=red]{hyperref}
\usepackage{orcidlink}

\newcommand\BibTeX{{\rmfamily B\kern-.05em \textsc{i\kern-.025em b}\kern-.08em
T\kern-.1667em\lower.7ex\hbox{E}\kern-.125emX}}

\def\volumeyear{2016}

\begin{document}

\runninghead{Jia et~al.}

\title{M2UD: A Multi-model, Multi-scenario, Uneven-terrain Dataset for Ground Robot with Localization and Mapping Evaluation}

\author{Yanpeng Jia\affilnum{1,2}~\orcidlink{0009-0008-1295-7439}, Shiyi Wang\affilnum{1}, Shiliang Shao\affilnum{1}~\orcidlink{0000-0002-4512-167X}, Yue Wang\affilnum{3}, Fu Zhang\affilnum{4} and Ting Wang\affilnum{1}~\orcidlink{0000-0003-2616-3150}}

\affiliation{\affilnum{1}State Key Laboratory of Robotics at Shenyang Institute of Automation, Chinese Academy of Sciences, Shenyang, China\\
\affilnum{2}University of Chinese Academy of Sciences, Beijing, China\\
\affilnum{3}Zhejiang University, Zhejiang, China\\
\affilnum{4}Department of Mechanical Engineering, The University of Hong Kong, Hong Kong}

\corrauth{Ting Wang and Shiliang Shao, State Key Laboratory of Robotics at Shenyang Institute of Automation, Chinese Academy of Sciences, No. 114, Nanta Street, Shenyang, 110016, China}
\email{wangting@sia.cn, shaoshiliang@sia.cn}

\begin{abstract}
As one of the most widely used types of robots, ground robots play a crucial role in inspection, exploration, rescue, and other applications. In recent years, advancements in Light Detection and Ranging (LiDAR) technology have made sensors more accurate, lightweight, and cost-effective. Therefore, researchers increasingly integrate LiDAR with other sensors, such as Inertial Measurement Units (IMUs) and cameras, for Simultaneous Localization and Mapping (SLAM) (SLAM) studies, providing robust technical support for ground robots and expanding their application domains. Public datasets that incorporate multiple sensors and diverse scenarios are essential for advancing SLAM technology in ground robots. However, existing datasets for ground robots are typically restricted to flat-terrain motion with 3 Degrees of Freedom (DOF) and cover only a limited range of scenarios. Although handheld devices and unmanned aerial vehicle (UAV) exhibit richer and more aggressive movements, their datasets are predominantly confined to small-scale environments due to endurance limitations. To fill these gap, we introduce M2UD, a multi-modal, multi-scenario, uneven-terrain SLAM dataset for ground robots. This dataset contains a diverse range of highly challenging environments, including cities, villages, open fields, long corridors, plazas, underground parking, and mixed scenarios. Additionally, it presents extreme weather conditions such as darkness, smoke, snow, and dust. The aggressive motion and degradation characteristics of this dataset not only pose challenges for testing and evaluating existing SLAM methods but also advance the development of more advanced SLAM algorithms. To benchmark SLAM algorithms, M2UD provides smoothed ground truth localization data obtained via Real-time Kinematics (RTK) and introduces a novel localization evaluation metric that considers both accuracy and efficiency. Additionally, we utilize a high-precision millimeter-level laser scanner to acquire ground truth maps of two representative scenes, facilitating the development and evaluation of mapping algorithms. We select 12 localization sequences and 2 mapping sequences to evaluate several classical LiDAR and visual SLAM algorithms, verifying usability of the dataset. To enhance usability, the dataset is accompanied by a suite of development kits, including data transformation, timestamp alignment, ground truth smooth, et. al. The dataset and related videos are available at \href{https://yaepiii.github.io/M2UD/}{https://yaepiii.github.io/M2UD/}.
\end{abstract}

\keywords{Dataset, ground robots, multi-sensor fusion, LiDAR, camera, Simultaneous Localization and Mapping (SLAM), Global Navigation Satellite System, Inertial Measurement Unit}

\maketitle

\section{Introduction}
Intelligent ground robots are widely deployed in various applications, including logistics, rescue, inspection, cleaning, and food distribution. In these scenarios, robots must operate reliably in indoor or mixed indoor-outdoor environments with Global Navigation Satellite System (GNSS)-deny. In recent years, as Light Detection and Ranging (LiDAR) technology has become more lightweight, cost-effective, and precise, multi-sensor fusion SLAM techniques (\cite{lvi-sam, fast-livo, r3live}) integrates 3D LiDAR with Inertial Measurement Units (IMUs), cameras, and GNSS, which have emerged as key enablers for the practical deployment of ground robots because them enable the construction of precise environmental maps while simultaneously tracking the robot’s position in real time. Despite significant advancements in Simultaneous Localization and Mapping (SLAM) research over the past few decades, many state-of-the-art solutions (\cite{fast-lio, lio-sam}) struggle to deploy on ground robots in typical scenarios. For example, visual SLAM often fails in dark or textureless environments; LiDAR SLAM suffers from degradation in structurally repetitive scenes such as long corridors and open areas; SLAM algorithms struggle in highly dynamic urban settings and crowded shopping malls; and LiDAR SLAM tends to accumulate vertical errors in uneven outdoor terrains. These failures prompt us to collect a SLAM dataset with more realistic, challenging scenarios to evaluate existing algorithms and advance the development of more robust SLAM algorithms.

\begin{figure*}[tbp]
	\centering
	\includegraphics[width=\textwidth]{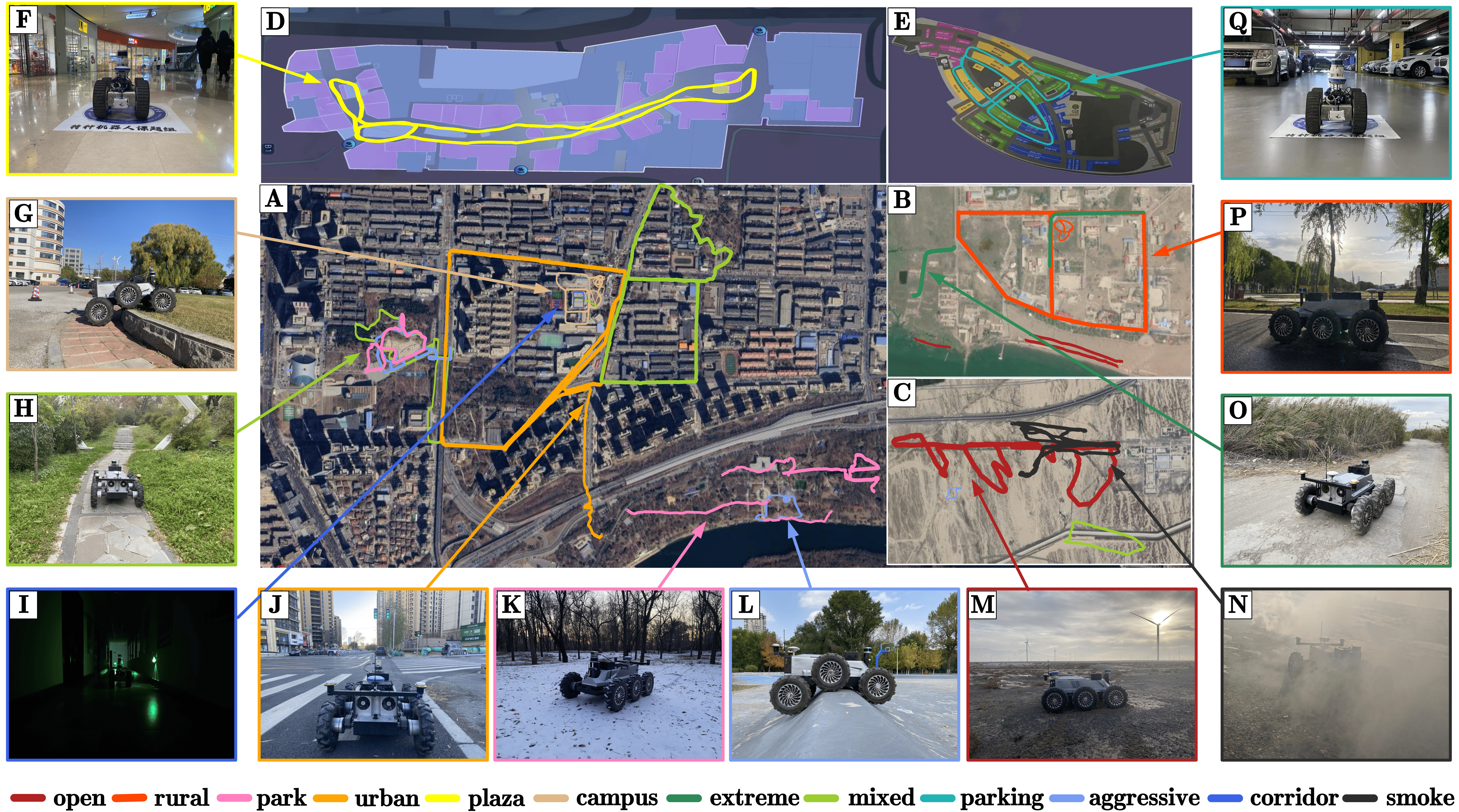}
	\caption{\textbf{Acquisition Zone trajectories.} We visualize the trajectories across 12 categories using distinct colors. (A)–(C) show satellite maps of our outdoor data recording zones; (D)–(E) present the floor plans of two indoor environments; (F)–(Q) provide snapshots of the robot and its surroundings during data collection, offering a visual representation of the recording conditions and robot status.}
	\label{figure1}
\end{figure*}

Available public datasets lowers the cost of SLAM technology development and establishes a fair benchmark for evaluating different algorithms. However, most existing ground robot datasets are restricted to 3-Degrees of Freedom (DOF) motion (\cite{darpa, kitti, kitti-360, oxford_robotcar, complex_urban}) and contain only a limited range of scenarios (\cite{ncd1, marulan, rosario, m2dgr}), which deviates from real-world applications. Although hand-held and unmanned aerial vehicle (UAV) datasets exhibit richer and more aggressive motions, which introduce challenges for SLAM, they do not align with the motion patterns of ground robots. Furthermore, due to endurance limitations, these datasets are typically restricted to small-scale environments (\cite{euroc_uav, uzh_fpv}) or lack LiDAR data (\cite{zurich_urban_mav}), making them unsuitable for large-scale, long-term applications (\cite{mars_livg}). To advance the application of SLAM in ground robots, we present a multi-modal, multi-scenario, uneven-terrain SLAM dataset, collected using two special ground robot platforms equipped with well-calibrated multi-channel LiDAR, RGB-D cameras, high-frequency IMUs, and GNSS, to be designed for maximum replicating the motion patterns of ground robots. Additionally, to reproduce various common application scenarios of ground robots, we collect a series of scenarios around the Shenyang Institute of Automation, Chinese Academy of Sciences (Figure~\ref{figure1} (A)) and in Xinjiang, China (Figure~\ref{figure1} (B), (C)). The dataset includes 58 sequences across 12 categories, collected from a diverse set of highly challenging environments, including urban areas, rural regions, open fields, long corridors, plazas, underground parking, and mixed scenarios, covering a total distance of over 50 km. Moreover, we collect data under extreme weather conditions, including darkness, smoke, snow, sand, and dust, to support researchers in developing SLAM solutions for such extreme situations. The data collection trajectory and scenario snapshots of this dataset are illustrated in Figure~\ref{figure1}.

To benchmark SLAM algorithms, the dataset provides smoothed location ground truth obtained via Real-time Kinematics (RTK) and ground truth maps of two representative scenes acquired using a high-precision millimeter-level laser scanner, facilitating the development and evaluation of localization and mapping algorithms. Additionally, current evaluations of SLAM algorithm localization performance primarily focus on accuracy comparisons, while in practical applications, algorithm efficiency is also a crucial consideration. Therefore, we propose a localization evaluation metric that considers both accuracy and efficiency, aiding developers in selecting algorithms better suited for real-world engineering applications. To enhance usability, the dataset is accompanied by a suite of development kits, including data transformation, timestamp alignment, ground truth smooth, etc.

In summary, we present a multi-sensor, multi-scene, uneven-terrain dataset for ground robots with localization and mapping evaluation, distinguished by the following five key characteristics:
\begin{itemize}[leftmargin=*]
\setlength{\itemsep}{0pt}
\setlength{\parsep}{0pt}
\setlength{\parskip}{0pt}
\item[$\bullet$] The dataset is collected using two special ground robots equipped with well-calibrated multi-channel LiDAR, an RGB-D camera, a high-frequency IMU, and GNSS, featuring 6-DOF aggressive motion characteristics not found in other ground robot datasets;
\item[$\bullet$] The dataset consists of 58 sequences categorized into 12 types, covering a total distance of over 50 km;
\item[$\bullet$] The dataset offers smoothed localization ground truth derived from RTK data and introduces a novel localization evaluation metric that accounts for both accuracy and efficiency.
\item[$\bullet$] This dataset offers ground truth maps of two representative scenes, acquired using a high-precision millimeter-level laser scanner, to support the development and evaluation of mapping algorithms.
\item[$\bullet$] The dataset is accompanied by a suite of development kits designed to enhance usability for developers, including data transformation, timestamp alignment, truth smoothing and etc.
\end{itemize}

The structure of this paper is as follows. Section 2 provides a detailed comparison between our dataset and existing datasets. A detailed description of the recording platform, sensor configuration, and calibration is presented in Section 3. Section 4 outlines the data recording scenarios and the characteristics of 12 categories, along with the localization and mapping ground truth, which also briefly introduces the development kits released with the dataset. In Section 5, we validate the dataset and present the results and analysis.

\section{Related work}
Table~\ref{table1} presents a selection of classic datasets, including those collected using handheld (\cite{tum_rgb-d, tum_vi, ncd2, vivid++, hilti-oxford}), aerial (\cite{euroc_uav, uzh_fpv, zurich_urban_mav, ntu_viral, mars_livg}), and Vehicle platforms (\cite{darpa, kitti, kitti-360, oxford_robotcar, complex_urban}). Table~\ref{table2} demonstrates that our dataset encompasses a broader range of scenarios, exhibits more intense motion, and features more uneven terrain compared to existing ground robot datasets \cite{rawseeds, marulan, nclt, m2dgr}.

\subsection{Other platform datasets}

\subsubsection{Handheld datasets:}
As one of the most widely used handheld datasets for visual SLAM applications, the TUM RGB-D dataset (\cite{tum_rgb-d}) captures handheld data from indoor office and corridor environments using Xbox Kinect depth cameras and introduces the widely recognized SLAM evaluation metrics: Absolute Trajectory Error (ATE) and Relative Pose Error (RPE). Their subsequent work (\cite{tum_vi}) further extends the dataset for monocular vision-inertial SLAM evaluation. However, due to the limited range of the motion capture system, their ground truths are only available near the beginning and end of cross-room sequences. Moreover, the datasets primarily focus on visual SLAM evaluation and does not include LiDAR data. With advancements in LiDAR technology, recent handheld datasets have included LiDAR sensors, such as the Newer College dataset (\cite{ncd2}), the ViViD++ dataset (\cite{vivid++}), and the Hilti-Oxford dataset (\cite{hilti-oxford}). Notably, the Newer College dataset (\cite{ncd2}) also provides ground truth for high-precision scene point clouds and mesh maps, serving as a benchmark for evaluating mapping algorithms. Due to the payload capacity constraints and limited mobility of handheld devices, most of handheld datasets primarily focus on relatively small areas, such as university campuses. Consequently, the evaluation of SLAM algorithms using such datasets is limited by scene coverage and scale.

\begin{table*}[!t]
	\caption{Several classic datasets, including handheld platforms, airborne platforms and ground robot platforms. I, L, C, and G indicate IMU, LiDAR, Camera, and GNSS respectively.}
	\centering
	\newcolumntype{C}[1]{>{\centering\arraybackslash}m{#1}<{\centering}}
	\scalebox{0.87}{
		\begin{tabular}{C{1.5cm}C{1.2cm}C{1.2cm}C{2cm}C{2cm}C{1.2cm}ccccC{1.8cm}c}
			\toprule
			\multirow{2}{*}{\textbf{Platform}} & \multirow{2}{*}{\textbf{Dataset}} & \multirow{2}{*}{\textbf{Year}} & \multirow{2}{*}{\textbf{Scenarios}} & \multirow{2}{*}{\textbf{Characteristics}} & \multirow{2}{*}{\textbf{Weather}}  & \multicolumn{4}{c}{\textbf{Sensors}} & \multirow{2}{*}{\textbf{\makecell{Localization \\ GroundTruth}}} & \multirow{2}{*}{\textbf{\makecell{Mapping \\ GroundTruth}}} \\
			\cline{7-10}
			\rule{-1pt}{10.5pt}
			& & & & & & \textbf{I} & \textbf{L} & \textbf{C} & \textbf{G} & & \\
			\midrule
			
			\multirow{5}{*}[-1.3cm]{\textbf{Handheld}} & TUM RGB-D & 2012 & Office/Hall & Structured & Sunny & \ding{51} & \ding{55} & \ding{51} & \ding{55} & 8*Mocap & \ding{55} \\
			\cline{2-12}
			\rule{-1pt}{10.5pt}
			
			& TUM VI & 2018 & Office/ Campus & Structured & Sunny & \ding{51} & \ding{55} & \ding{51} & \ding{55} & Mocap & \ding{55} \\
			\cline{2-12}
			\rule{-1pt}{10.5pt}
			
			& Newer College & 2021 & Park/ Campus & Structured/ Vegetated & Sunny & \ding{51} & \ding{51} & \ding{51} & \ding{55} & 6-DOF ICP & \ding{51} \\
			\cline{2-12}
			\rule{-1pt}{10.5pt}
			
			& ViViD++ & 2022 & Campus/ Urban & Structured & Changing light & \ding{51} & \ding{51} & \ding{51} & \ding{51} & SLAM/ RTK & \ding{55} \\
			\cline{2-12}
			\rule{-1pt}{10.5pt}
			
			& Hilti-Oxford & 2023 & Construction site & Structured/ Multilevel & Changing light & \ding{51} & \ding{51} & \ding{51} & \ding{51} & 6-DOF ICP & \ding{51} \\
			\midrule
			
			\multirow{5}{*}[-1.6cm]{\textbf{UAV}} & EuRoC MAV & 2016 & Hall & Unstructured & Sunny & \ding{51} & \ding{55} & \ding{51} & \ding{55} & Mocap & \ding{51} \\
			\cline{2-12}
			\rule{-1pt}{10.5pt}
			
			& Zurich Urban MAV & 2017 & Urban & Structured/ Dynamic & Sunny & \ding{51} & \ding{55} & \ding{51} & \ding{55} & Aerial Photogrammetry & \ding{55} \\
			\cline{2-12}
			\rule{-1pt}{10.5pt}
			
			& UZH-FPV & 2019 & Grassland & Unstructured & Sunny & \ding{51} & \ding{55} & \ding{51} & \ding{55} & Laser Tracker & \ding{55} \\
			\cline{2-12}
			\rule{-1pt}{10.5pt}
			
			& NTU VIRAL & 2022 & Campus & Structured & Sunny & \ding{51} & \ding{51} & \ding{51} & \ding{55} & Laser Tracker & \ding{55} \\
			\cline{2-12}
			\rule{-1pt}{10.5pt}
			
			& MARS- LIVG & 2024 & Field/ Rural & Unstructured/ Higher altitudes & Sunny & \ding{51} & \ding{51} & \ding{51} & \ding{51} & RTK & \ding{51} \\
			\midrule
			
			\multirow{8}{*}[-2cm]{\textbf{Vehicle}} & DARPA & 2010 & Urban & Structured/ Jam & Sunny & \ding{55} & \ding{51} & \ding{51} & \ding{51} & DGPS/INS & \ding{55} \\
			\cline{2-12}
			\rule{-1pt}{10.5pt}
			
			& KITTI & 2013 & Urban/ Rural & Structured & Sunny & \ding{51} & \ding{51} & \ding{51} & \ding{51} & RTK/INS & \ding{55} \\
			\cline{2-12}
			\rule{-1pt}{10.5pt}
			
			& Oxford RobotCar & 2017 & Urban & Structured/ Long-term & Sunny/ Rain/ Snow & \ding{51} & \ding{51} & \ding{51} & \ding{51} & GPS/INS Not recommended & \ding{55} \\
			\cline{2-12}
			\rule{-1pt}{10.5pt}
			
			& Apollo Scape & 2018 & Urban & Structured & Sunny/ Rain & \ding{51} & \ding{51} & \ding{51} & \ding{51} & GPS/INS & \ding{55} \\
			\cline{2-12}
			\rule{-1pt}{10.5pt}
			
			& Complex Urban & 2019 & Urban & Structured/ Diverse urban & Sunny & \ding{51} & \ding{51} & \ding{51} & \ding{51} & SLAM & \ding{51} \\
			\cline{2-12}
			\rule{-1pt}{10.5pt}
			
			& Semantic KITTI & 2019 & Urban/ Rural & Structured & Sunny & \ding{51} & \ding{51} & \ding{51} & \ding{51} & RTK/INS & \ding{55} \\
			\cline{2-12}
			\rule{-1pt}{10.5pt}
			
			& UrbanLoco & 2020 & Urban & Structured/ Dynamic & Sunny & \ding{51} & \ding{51} & \ding{51} & \ding{51} & GPS/INS & \ding{55} \\
			\cline{2-12}
			\rule{-1pt}{10.5pt}
			
			& KITTI-360 & 2023 & Urban & Structured/ Dynamic & Sunny & \ding{55} & \ding{51} & \ding{51} & \ding{51} & GPS/INS & \ding{55} \\
			\midrule
			
			\textbf{Ground Robot} & Ours & 2025 & Urban/Campus /Field/Rural Open/Corridor & Unstructured/ Degraded/ Dynamic/ Aggressive & Sunny/ Snow/ Dust/ Smoke & \ding{51} & \ding{51} & \ding{51} & \ding{51} & RTK/SLAM & \ding{51} \\
			
			\bottomrule
		\end{tabular}
	}
	\label{table1}
\end{table*}

\subsubsection{Aerial datasets:}
UAVs equipped with various sensors offer a unique aerial perspective for SLAM. Due to safety considerations, early aerial datasets, such as EuRoC MAV (\cite{euroc_uav}) and UZH-FPV Drone Racing (\cite{uzh_fpv}), primarily focus on indoor or small-scale environments. With advancements in UAV endurance and safety, the Zurich Urban MAV dataset (\cite{zurich_urban_mav}) has extended aerial data collection to urban environments. However, due to payload capacity constraints, these UAVs do not carry 3D LiDAR and are only suitable for evaluating visual SLAM algorithms. With the growing demand for LiDAR SLAM and lightweight advancements in LiDAR, recent aerial datasets (\cite{ntu_viral, mars_livg}) are increasingly including rotational or solid-state LiDAR for data collection while gradually expanding to larger-scale environments to support multi-sensor fusion and wide-area operations. Although aerial datasets exhibit more aggressive and diverse motion patterns, which introduce new challenges for SLAM algorithms, most aerial datasets include only a limited set of sensors and are restricted to small-scale, short-duration environments, similar to handheld datasets. Furthermore, due to safety concerns, UAVs cannot operate in densely populated areas, such as plazas or underground parking, nor can they provide data sequences under extreme weather conditions.

\subsubsection{Vehicle-platform datasets:}
The datasets primarily focus on urban environments, for instance, the DARPA dataset (\cite{darpa}), originalating from the Urban Challenge competition, captures complex urban scenarios, including high-speed driving, sharp turns, traffic congestion, and vehicle collisions. Another highly representative dataset is the KITTI dataset (\cite{kitti}), one of the most widely used datasets in SLAM and autonomous driving research. It provides raw data collected from a Velodyne HDL-64E LiDAR, two color PointGrey Flea2 cameras, and two grayscale PointGrey Flea2 cameras, while generating smoothed trajectory ground truth using RTK/INS. However, the KITTI dataset provides only low-frequency IMU data, preventing the development of tightly coupled LiDAR/visual SLAM algorithms on this platform. Furthermore, the KITTI dataset is relatively dated, as it is collected more than a decade ago, whose data collection conditions are highly ideal, with favorable lighting and weather, rich textures and structures, and predominantly static scenes, leading to an impressive result of 0.53\% error in translation (\cite{loam}). This motivates the presentation of Semantic KITTI (\cite{semantic_kitti}) and KITTI-360 (\cite{kitti-360}), which adhere to the data specifications of the KITTI dataset, while introducing more challenging scenarios, and extending the KITTI series for advanced semantic segmentation and object recognition. The Oxford RobotCar dataset (\cite{oxford_robotcar}) offers long-term (over 1,000 km, over 1 year) LiDAR and camera data collected. Over this period, it captures road and large-scale building construction and modifications, as well as typical weather conditions relevant to autonomous driving, such as heavy rain and snow, which has significantly contributed to research on long-term SLAM and scene recognition. However, its trajectory ground truth is only accurate within a few hundred meters. The Complex Urban (\cite{complex_urban}), UrbanLoco (\cite{urbanloco}), and ApolloScape (\cite{apolloscape}) datasets further expand the scale and urban complexity of vehicle-platform data, incorporating more challenging high-dynamic scenarios and cross-city sequences, which introduce new challenges for SLAM development. In conclusion, most vehicle-platform datasets are primarily designed for autonomous driving applications in urban or rural environments. The sensor configurations are relatively ideal and provide only 3-DOF flat-ground motion, which masks the limitations of SLAM algorithms in real-world applications, particularly in more challenging scenarios and with lower-cost sensors.

\subsection{Ground robot datasets}

\begin{table*}[!t]
	\caption{Comparison of ground robots datasets. I, L, C, and G indicate IMU, LiDAR, Camera, and GNSS respectively. The Maximum Accelerometer  Variance (MAV) and Maximum Gyroscope Variance (MGV) quantify the motion intensity. TD and SLD denote the Total Distance and the Single Longest Distance, respectively. The Local Maximum Elevation Difference (LMED) characterizes terrain unevenness. The best value for each metric is highlighted in \textbf{bold}, while the second-best value is \underline{underlined}.}
	\centering
	\newcolumntype{C}[1]{>{\centering\arraybackslash}m{#1}<{\centering}}
	\scalebox{0.82}{
		\begin{tabular}{C{1.2cm}C{1cm}C{1.2cm}C{1.2cm}C{1.2cm}cccccccccc}
			\toprule
			\multirow{2}{*}{\textbf{Dataset}} & \multirow{2}{*}{\textbf{Year}} & \multirow{2}{*}{\textbf{\makecell{Scenarios \\ Number}}} & \multirow{2}{*}{\textbf{\makecell{Extreme \\ Weather}}} & \multirow{2}{*}{\textbf{Degraded}} & \multicolumn{4}{c}{\textbf{Sensors}} & \multirow{2}{*}{\textbf{\makecell{Mapping \\ GroundTruth}}} & \multirow{2}{*}{\textbf{MAV}} & \multirow{2}{*}{\textbf{MGV}} & \multirow{2}{*}{\textbf{TD}} & \multirow{2}{*}{\textbf{SLD}} & \multirow{2}{*}{\textbf{LMED}} \\
			\cline{6-9}
			\rule{-1pt}{10.5pt}
			& & & & & \textbf{I} & \textbf{L} & \textbf{C} & \textbf{G} & & & & & & \\
			\midrule
			
			Rawseeds & 2009 & 1 & \ding{55} & \ding{55} & \ding{51} & \ding{51} & \ding{51} & \ding{51}  & \ding{55} & 5.97 & 0.12 & 12295.91 & 2263.95 & 0.01 \\
			\midrule
			
			New College & 2009 & 2 & \ding{55} & \ding{55} & \ding{51} & \ding{51} & \ding{51} & \ding{51} & \ding{51} & 4.46 & 0.15 & 5847.52 & 225.11 & \underline{7.88} \\
			\midrule
			
			Marulan & 2010 & 1 & Artificial & \ding{55} & \ding{51} & \ding{51} & \ding{51} & \ding{51} & \ding{55} & 1.16 & 0.06 & 1421.00 & 193.06 & 1.71 \\
			\midrule
			
			NCLT & 2016 & 1 & \ding{55} & \ding{55} & \ding{51} & \ding{51} & \ding{51} & \ding{51} & \ding{55} & 0.34 & 0.88 & \textbf{147432.54} & \textbf{7582.54} & 3.69 \\
			\midrule
			
			Chilean & 2017 & 1 &\ding{55} & \ding{51} & \ding{55} & \ding{51} & \ding{51} & \ding{55} & \ding{55} & - & - & 1841.33 & 1841.33 & - \\
			\midrule
			
			Rosario & 2019 & 1 & \ding{55} & \ding{55} & \ding{51} & \ding{55} & \ding{51} & \ding{51} & \ding{55} & \underline{20.74} & 0.12 & 1970.42 & 529.14 & 1.49 \\
			\midrule
			
			Open-LORIS & 2020 & \underline{4} & \ding{55} & \ding{55} & \ding{51} & \ding{51} & \ding{51} & \ding{55} & \ding{55} & 2.03 & 0.14 & 1503.08 & 230.30 & 1.10\\
			\midrule
			
			NAVER LABS & 2021 & 1 & \ding{55} & \ding{55} & \ding{55} & \ding{51} & \ding{51} & \ding{55} & \ding{51} & - & - & 3397.30 & 1181.66 & 4.86 \\
			\midrule
			
			M2DGR & 2022 & 3 & \ding{55} & \ding{55} & \ding{51} & \ding{51} & \ding{51} & \ding{51} & \ding{55} & 2.64 & 0.57 & 10708.67 & 1484.62 & 6.13 \\
			\midrule
			
			MAgro dataset & 2023 & 1 & \ding{55} & \ding{55} & \ding{51} & \ding{51} & \ding{51} & \ding{51} & \ding{55} & - & - & - & - & - \\
			\midrule
			
			Under canopy & 2024 & 1 & \ding{55} & \ding{55} & \ding{51} & \ding{55} & \ding{51} & \ding{51} & \ding{55} & 10.66 & 0.86 & 976.99 & 284.07 & 0.09 \\
			\midrule
			
			CID-SIMS & 2025 & 2 & \ding{55} & \ding{55} & \ding{51} & \ding{55} & \ding{51} & \ding{55} & \ding{51} & 0.36 & \underline{3.84} & 2134.21 & 196.45 & 0.66 \\
			\midrule
			
			Ours & 2025 & \textbf{8} & \ding{51} & \ding{51} & \ding{51} & \ding{51} & \ding{51} & \ding{51} & \ding{51} & \textbf{44.93} & \textbf{10.49} & \underline{51008.32} & \underline{6350.14} & \textbf{12.68} \\
			
			\bottomrule
		\end{tabular}
 	}
	\label{table2}
\end{table*}

\subsubsection{Existing datasets:}
Owing to the superior payload capacity and endurance compared to UAVs and handheld devices, ground robots are typically equipped with a more comprehensive suite of sensors. Rawseeds (\cite{rawseeds}) and New College (\cite{ncd1}) are among the earliest ground robot datasets, while the immaturity of 3D LiDAR technology at the time, both datasets use only 2D LiDARs and are limited to small-scale scene collection. The Marulan dataset (\cite{marulan}) captures both day and night scenes in rural areas and artificially simulates various weather conditions, including sandstorms, smoke, and rainfall, thereby introducing unstructured characteristics into the ground robot dataset. NCLT (\cite{nclt}) is the first long-term dataset for ground robots, which is equipped with a 3D LiDAR and performs campus explorations approximately every two weeks over 15 months, capturing data across different times of the day, both indoors and outdoors, throughout all four seasons. However, the scenarios in the NCLT dataset remain confined to the campus environment. The Chilean Underground Mine dataset (\cite{chilean}) expands the application of ground robots to subterranean environments, which provides precise 6-DOF ground-truth poses through employing the ICP algorithm for offline matching using a survey-grade LiDAR scanner. The Rosario dataset (\cite{rosario}) focuses on the application of ground robots in agriculture and captures farmland data using a ZED stereo camera. The recent Under-canopy dataset (\cite{under-canopy}) further extends the use of ground robots in agriculture, collecting visual-inertial data over a three-month period during the growth of corn and soybeans. Both datasets exhibit LiDAR-degradeds and overexposure due to direct sunlight, posing challenges to visual SLAM algorithms, but neither dataset includes LiDAR data. The MAgro dataset (\cite{magro}) employs the same camera as the two aforementioned datasets while additionally integrating 3D LiDAR data. However, its nine sequences are confined to just two paths, with no path over 500 meters. The OpenLORIS-Scene dataset (\cite{openLORIS}) collects LiDAR, IMU, and RGB-D data across various indoor environments, supporting the validation of SLAM in service, cleaning, as well as the development of lifelong algorithms. The NAVER LABS dataset (\cite{naver_labs}) primarily captures scenes in shopping malls, characterized by dynamic crowds and varying appearances, while it is relatively small in scale and lacks IMU data. M2DGR (\cite{m2dgr}) is a recently introduced ground robot dataset capturing multi-sensor data from a campus and its surrounding streets, incorporating infrared and event cameras, which are gaining increasing attention. However, the robot motion in M2DGR is relatively smooth and slow, with most sequences recorded in small-scale areas (the longest sequence lasts 1,227 seconds and does not exceed 2 km). CID-SIMS (\cite{cid-sims}) presents a dataset for SLAM and 3D reconstruction in indoor environments, offering image semantic annotations. However, its scene scale is relatively limited, and it unsuitable for LiDAR SLAM algorithms. In summary, existing ground robot datasets encompass a limited range of scene types and primarily feature smooth 3-DOF motion. Also, due to safety concerns, these datasets include only a limited number of extreme weather conditions. As ground robots are applied to increasingly diverse domains, there is a growing need for more challenging real-world datasets featuring complex motion to advance related algorithm development.

\begin{figure*}[!tbp]
	\centering
	\includegraphics[width=\textwidth]{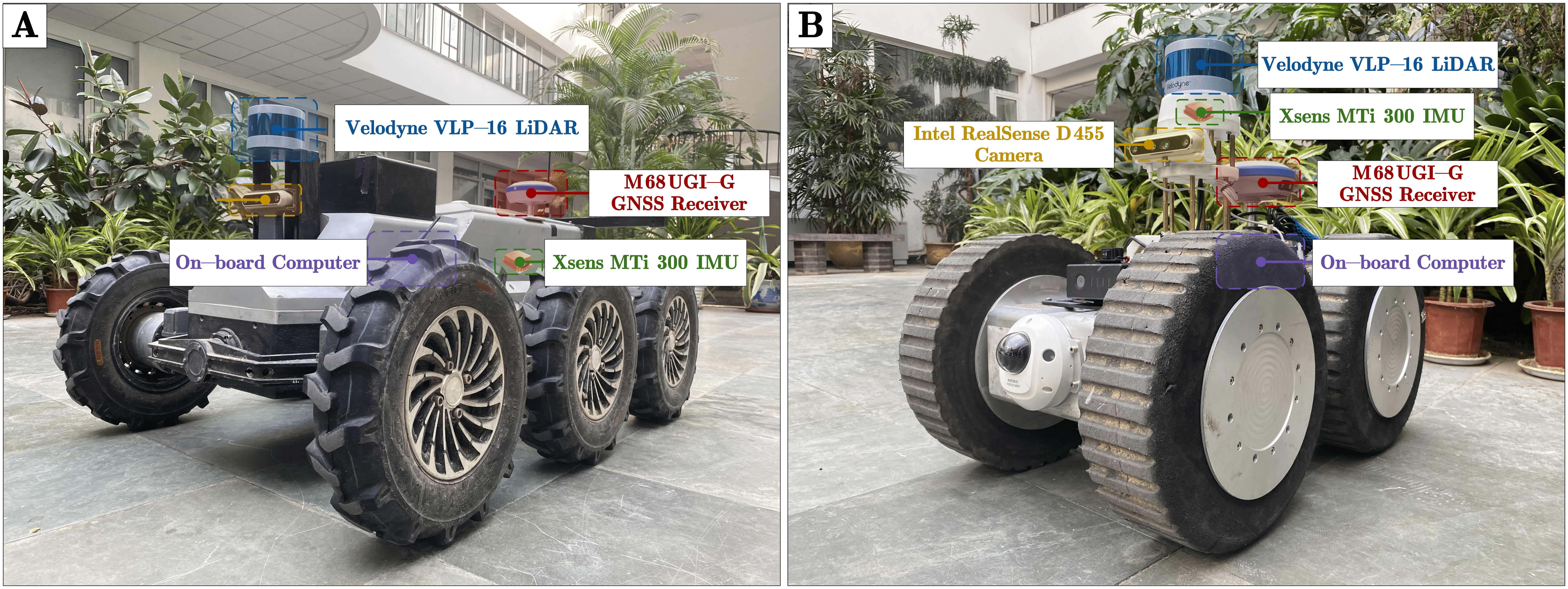}
	\caption{\textbf{Acquisition Platforms.} (A) Six-wheeled special robot primarily used for outdoor data collection. (B) Four-wheeled mobile robot primarily used for indoor data collection.}
	\label{figure2}
\end{figure*}

\subsubsection{The proposed dataset:}
The proposed M2UD dataset seeks to bridge the gap in existing ground robot datasets and enhance the application of ground robot SLAM across various domains. Specifically, compared to the Rawseeds (\cite{rawseeds}) and New College (\cite{ncd1}) datasets, M2UD offers more comprehensive sensor data, better suited for the advancement of modern SLAM algorithms. Unlike Marulan (\cite{marulan}), M2UD captures real extreme weather conditions over nearly a year, including darkness (Figure~\ref{figure1} (I)), snow (Figure~\ref{figure1} (K)), sandstorms (Figure~\ref{figure1} (O)), and smoke (Figure~\ref{figure1} (N)), among others. Datasets such as Rosario (\cite{rosario}), Under-canopy (\cite{under-canopy}), OpenLORIS-Scene (\cite{openLORIS}), and NAVER LABS (\cite{naver_labs}) primarily focus on specific scenarios. In contrast, M2UD consist of 12 categories with 58 sequences, covering 8 different scenarios, such as urban areas (Figure~\ref{figure1} (J)), rural areas (Figure~\ref{figure1} (P)), open field (Figure~\ref{figure1} (M)), parks (Figure~\ref{figure1} (K)), campus (Figure~\ref{figure1} (G)), plazas (Figure~\ref{figure1} (F)), parking (Figure~\ref{figure1} (Q)), and mixed scenarios (Figure~\ref{figure1} (H)). The dataset covers over 50 km, with the longest single sequence extending up to 6.3 km, making it the second longest after the Long-Term dataset NCLT \cite{nclt}. However, it is crucial to note that NCLT \cite{nclt} consists solely of repeatedly recorded campus routes. Additionally, as shown in Figure~\ref{figure1}, these sequences also include LiDAR-degraded scenarios, such as dark long corridors (Figure~\ref{figure1} (I)) and open areas (Figure~\ref{figure1} (M)), as well as visual-degraded scenarios, including highly-dynamic environments (Figure~\ref{figure1} (J)) and occlusion (Figure~\ref{figure1} (N)), which is uncommon in other ground robot datasets. Moreover, to quantitatively assess the motion intensity and terrain roughness, we define three metrics: Maximum Accelerometer  Variance (MAV), Maximum Gyroscope Variance (MGV), and Local Maximum Elevation Difference (LMED). For MAV and MGV, each sequence is divided into 50 s segments, the variance of acceleration and angular velocity is computed for each segment, and the maximum value across all segments is taken; For LMED, the elevation difference is computed every 20 m with a 10 m sliding step, and the maximum value is taken. Compared to all existing ground robot datasets, M2UD exhibits the highest MAV and MGV, as well as a significantly higher LMED than other datasets, which highlights that it features intense motion and highly uneven terrain, posing new challenges for SLAM. Finally, compared with the existing ground robot datasets (\cite{rawseeds, ncd1, marulan, rosario, naver_labs}), M2UD captures multi-model data under aggressive and uneven terrain conditions (Figure~\ref{figure1} (L)), exhibiting 6-DOF motion characteristics similar to those of handheld devices or UAV platforms.

\section{System overview}

\subsection{Acquisition platform}
As shown in Figure~\ref{figure2}, two self-developed ground robots are utilized for data collection, both equipped with identical sensors to ensure data consistency. The six-wheeled special robot in Figure~\ref{figure2} (A) is designed for outdoor data collection, whose unique rocker-arm design provides excellent adaptability to normal road surfaces (Figure~\ref{figure1} (J)), uneven terrains (Figure~\ref{figure1} (L)), and stairs (Figure~\ref{figure1} (G)), making it particularly suitable for challenging outdoor environments. As shown in Figure~\ref{figure2} (B), the four-wheeled mobile robot is compact, highly maneuverable, and capable of fast movement, which can navigate freely through various indoor environments and is primarily used for data collection in locations such as plazas (Figure~\ref{figure1} (F)), parking (Figure~\ref{figure1} (Q)), and long corridors (Figure~\ref{figure1} (I)). The Robot Operating System (ROS) is utilized to store data in bag format on a 4TB solid-state drive in the on-board computer, facilitating its use for SLAM algorithms.

\begin{figure*}[!tbp]
	\centering
	\includegraphics[width=\textwidth]{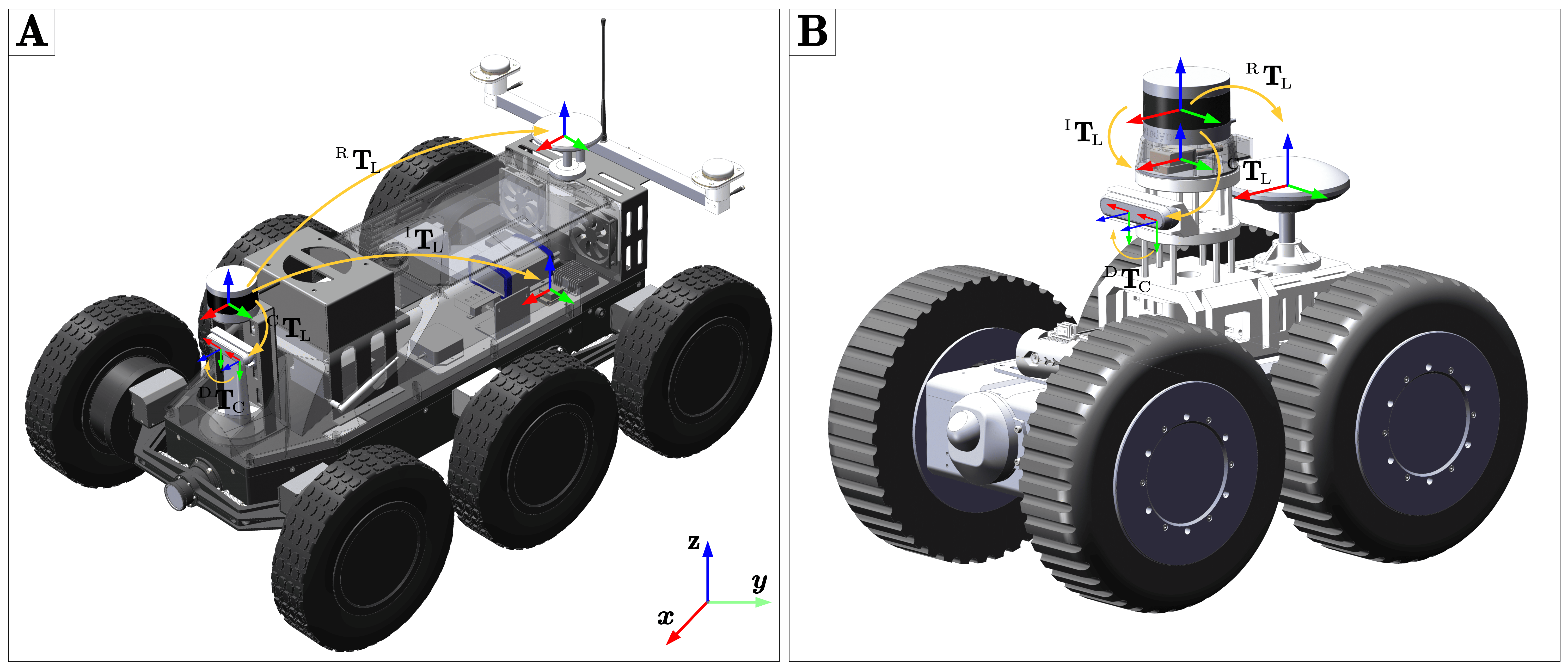}
	\caption{\textbf{The Coordinate System for the Acquisition platform.}}
	\label{figure3}
\end{figure*}

\subsection{Sensor setup}
We argue that while smooth motion, clear images, and dense point clouds obviously facilitate algorithm processing, they may not be sufficient for practical applications (\cite{survey}). Therefore, we select cost-effective and widely used sensors to replicate real-world robotic application conditions as closely as possible. In this section, we provide a detailed description of our sensor setup and parameters.
\begin{enumerate}
\item[(i)] \textbf{LiDAR:} We utilize the Velodyne VLP-16\endnote{https://ouster.com/products/hardware/vlp-16} LiDAR to capture 3D point clouds of the surrounding environment at 10 Hz, collecting over 300,000 points per second. The sensor features a 360$\degree$ horizontal field of view (FoV) and a 30$\degree$ vertical FoV, with a scanning range over 100 m. To ensure efficient data transmission, we connect the LiDAR to the host via an Ethernet port.

\item[(ii)] \textbf{Camera:} The Intel RealSense D455\endnote{https://www.intelrealsense.com/depth-camera-d455/} is a global shutter RGB-D camera capable of capturing time-synchronized color and depth images at a resolution of 720×480 pixels at 10 Hz, whose optimal depth range extends from 0.6 m to 6 m from the image plane, with an error margin of less than 2\% within a 4 m range.

\item[(iii)] \textbf{IMU:} To ensure higher accuracy in IMU data acquisition, we select not to use the internal IMU (Bosch BMI055) of the RealSense D455. Instead, we integrate the Xsens MTi 300\endnote{https://www.xsens.com/hubfs/Downloads/Leaflets/MTi-300.pdf} to capture 9-axis measurement data at 200 Hz. The accelerometer exhibits a bias stability of 15 $\mu g$ and a noise density of 60 $\mu g/\sqrt{Hz}$, the gyroscope achieves a bias stability of 10$\degree /s$ with a noise density of 0.003$\degree /s$, and the magnetometer provides a resolution of 0.25 $mG$.

\item[(iv)] \textbf{GNSS:} The M68UGI-G GNSS\endnote{https://www.devecent.com/M68.html} receiver is utilized to acquire raw positioning data, including GPS satellite count, timestamps, signal quality, and other relevant information at 5 Hz. By leveraging RTK technology, high-precision robot positions are estimated as ground truth for the trajectory, achieving a horizontal accuracy of $\pm(8mm+1ppm)$ and a vertical accuracy of $\pm(15mm+1ppm)$.
\end{enumerate}

\subsection{Calibration and time synchronization}
\begin{figure}[!tbp]
	\centering
	\includegraphics[width=0.5\textwidth]{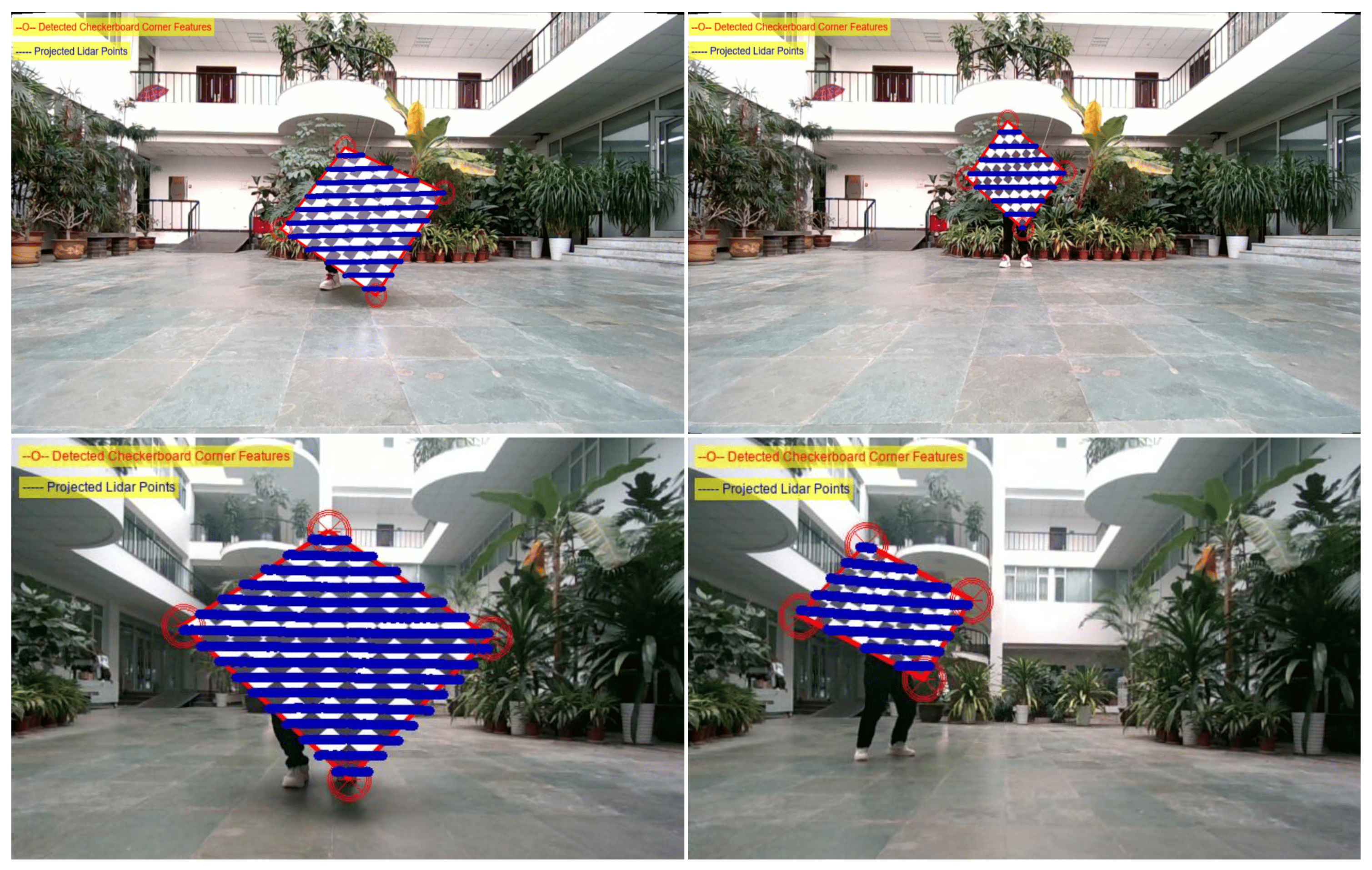}
	\caption{\textbf{Snapshot of the LiDAR-Camera Calibration Procedure.} The first row presents a snapshot of the calibration procedure for the six-wheel robot platform, while the second row displays a snapshot of the calibration procedure for the four-wheel robot platform.}
	\label{figure4}
\end{figure}

Our dataset offers precise intrinsic and extrinsic calibration parameters, and the coordinate systems of the sensors in both robots as shown in Figure~\ref{figure3}. This section presents the calibration methods and procedures. The parameters for all sequences are generated using the development kits in Section 4.3 and stored in \verb+calibration.yaml+.
\begin{enumerate}
	\item[(i)] \textbf{Intrinsic calibration:} For the camera intrinsic parameters, we use the manufacturer-provided intrinsic parameters for the color and depth cameras. Then, we collect a set of calibration images with a self-designed calibration board for validating the accuracy of these parameters using the MATLAB Camera Toolbox. The IMU intrinsic parameters are obtained by recording a static IMU sequence for over six hours at 200 Hz. The intrinsic parameters, including noise density and random walk bias, are calibrated using the imu\_utils\endnote{https://github.com/gaowenliang/imu\_utils} toolbox. To support the validation of other calibration algorithms, the calibration sequences are publicly available alongside the dataset.
	
	\item[(ii)] \textbf{Extrinsic calibration:} For the calibration of extrinsic parameters, an initial estimation is first obtained from the SolidWorks model. The extrinsic parameters between the LiDAR and IMU are obtained using the method described in \cite{lidar-imu-calib}; The Kalibr toolbox (\cite{camera-imu-calib}) is used to calibrate the extrinsic parameters between the camera and IMU; The extrinsic parameters between the color and depth cameras are provided by the manufacturer; The transformation between the LiDAR and camera is determined using the MATLAB LiDAR-Camera Calibration Toolbox. Although the sparsity of the LiDAR point cloud poses a challenge for calibration, snapshots of the calibration process (Figure~\ref{figure4}) demonstrate that the LiDAR point cloud is accurately projected onto the calibration plate, highlighting the precision of our extrinsic calibration. The originalal calibration data is publicly available on the dataset website.
	
	\item[(iii)] \textbf{Time synchronization:} Instead of using hardware signals to trigger all sensors simultaneously, we record data from different sensors with the same system timestamp. In other word, LiDAR, camera, and IMU are synchronized via software by calling the APIs to trigger data capture at the same instance (\cite{m2dgr, ncd2}). Test results indicate that this software synchronization method achieves a time synchronization accuracy of less than 10 ms.
\end{enumerate}

\begin{figure*}[!tbp]
	\centering
	\includegraphics[width=\textwidth]{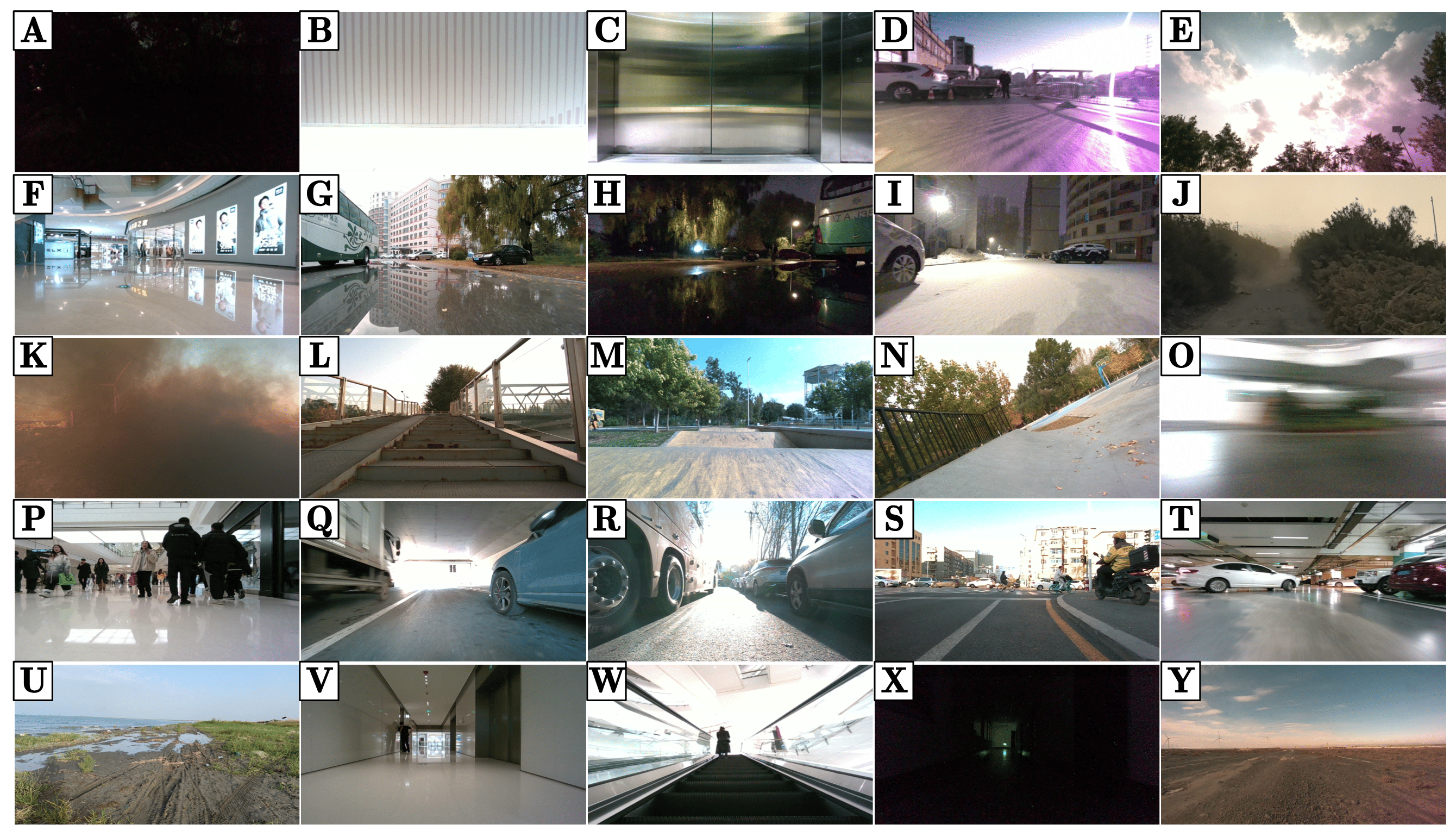}
	\caption{\textbf{The First-person Snapshots during The Collection.} (A)–(H) illustrate the impact on vision; (I)–(K) depict extreme weather conditions; (L)–(O) highlight instances of uneven-terrain and aggressive motion; (P)–(T) show highly-dynamic scenes; (U)–(Y) demonstrate LiDAR degradation elements within the dataset.}
	\label{figure5}
\end{figure*}

\subsection{Data format}
Table~\ref{table3} presents the information included in the ROS bags, along with their corresponding message types and frequencies. We aim to provide concise and comprehensive topics. For RTK data, the ROS bags record measurement positions, connection status, and GPS satellite conditions in the \verb+NavSatFix+ format. LiDAR data is provided in the widely used \verb+PointCloud2+ format to enhance usability. IMU data includes acceleration, angular velocity, and magnetometer readings. To preserve image quality, image data remains uncompressed, and the original image topic is provided as \verb+sensor_msgs/Image+. All ROS drivers for LiDAR, IMU, and cameras utilize manufacturer-released ROS drivers, while the RTK topic conversion script is open-sourced alongside our dataset.

\begin{table}[h]
	\caption{Data topic format information.}
	\centering
	\newcolumntype{C}[1]{>{\centering\arraybackslash}m{#1}<{\centering}}
	\scalebox{0.8}{
		\begin{tabular}{llll}
			\toprule
			\textbf{Type}   & \textbf{Topic Name}     & \textbf{Message Type}    & \textbf{Rate (Hz)} \\
			\midrule
			RTK    & /RTK/data             & sensor\_msgs/NavSatFix        & 5         \\
			LiDAR  & /velodyne\_points     & sensor\_msgs/PointCloud2      & 10        \\
			\multirow{2}{*}{IMU}    & /imu/data             & sensor\_msgs/Imu              & 200       \\
			& /imu/mag              & sensor\_msgs/MagneticField    & 100       \\
			\multirow{2}{*}{Camera} & /camera/color         & sensor\_msgs/Image            & 10        \\
			& /camera/depth         & sensor\_msgs/Image            & 10        \\
			\bottomrule
		\end{tabular}
	}
	\label{table3}
\end{table}

\section{Dataset}

\subsection{Sequence characteristics}
Our dataset comprises 58 sequences with specifically designed routes, categorized into 12 types based on collection scenarios or characteristics, with difficulty levels assigned for user reference. Figure~\ref{figure5} presents a view of the robot during data recording, illustrating that the sequences contain common application scenarios of ground robots as well as challenging conditions, including complete darkness (Figure~\ref{figure5} (A)), textureless (Figure~\ref{figure5} (B)), lift (Figure~\ref{figure5} (C)), overexposure (Figure~\ref{figure5} (D), (E)), reflections (Figure~\ref{figure5} (F)–(H)), extreme weather conditions (Figure~\ref{figure5} (I)–(K)), aggressive motion (Figure~\ref{figure5} (L)–(O)), dynamic objects (Figure~\ref{figure5} (P)–(T)), and open areas or highly-repetitive structures (Figure~\ref{figure5} (U)–(Y)). Our dataset offers trajectories and two maps ground truth, which can be directly utilized for evaluating localization and mapping in SLAM. The details of the 12 categories are presented in Table~\ref{table4} and detailed descriptions of each sequence are available on our dataset website.
\begin{table*}[!t]
	\caption{Statistical information and brief descriptions of 12 categories.}
	\centering
	\newcolumntype{C}[1]{>{\centering\arraybackslash}m{#1}}
	\newcolumntype{L}[1]{>{\raggedright\arraybackslash}m{#1}}
	\scalebox{0.8}{
		\begin{tabular}{C{1cm}C{1cm}C{1.4cm}C{2.5cm}C{3cm}C{2.5cm}L{7cm}}
			\toprule
			\textbf{Scene} & \textbf{Number} & \textbf{Size (GB)} & \textbf{Total Time (s)} & \textbf{Total Distance (m)} & \textbf{Average Difficulty} & \textbf{Description} \\
			\midrule
			open       & 8  & 60   & 5055.7  & 5596.7  & \textbf{\ding{72}\ding{72}\ding{73}\ding{73}\ding{73}} & Lakeside and open field, sparse features, no dynamics, uneven-terrain. \\
			rural      & 2  & 6.32 & 3111.9  & 3835.5  & \textbf{\ding{72}\ding{72}\ding{72}\ding{73}\ding{73}} & Unstable features, sparsely populated, flat-terrain. \\
			urban      & 5  & 100  & 6577.5  & 7770.0  & \textbf{\ding{72}\ding{72}\ding{72}\ding{72}\ding{73}} & High dynamics, various types of vehicles, flat-terrain. \\
			aggressive & 7  & 38.8 & 2075.8  & 2080.9  & \textbf{\ding{72}\ding{72}\ding{72}\ding{73}\ding{73}} & Continuous aggressive motion, uneven-terrain. \\
			campus     & 6  & 51.3 & 2728.2  & 3152.3  & \textbf{\ding{72}\ding{72}\ding{73}\ding{73}\ding{73}} & Research institute campus, structured environment, few dynamic objects, flat-terrain. \\
			park       & 6  & 61.7 & 2904.6  & 3205.2  & \textbf{\ding{72}\ding{72}\ding{73}\ding{73}\ding{73}} & Unstructured environment, few dynamic objects, uneven-terrain. \\
			extreme    & 4  & 0.97 & 1669.5  & 1976.8  & \textbf{\ding{72}\ding{72}\ding{72}\ding{73}\ding{73}} & Severe dust, snow, few dynamics, flat-terrain. \\
			mixed      & 5  & 116  & 9659.8  & 12161.7 & \textbf{\ding{72}\ding{72}\ding{72}\ding{72}\ding{72}} & Mixed scenarios such as indoor+outdoor, urban+park, parking+plaza, featuring characteristics of the mixed categories. \\
			smoke      & 4  & 16.8 & 1155.5  & 1164.9  & \textbf{\ding{72}\ding{72}\ding{72}\ding{73}\ding{73}} & Navigating through smoke, visual occlusion, open scenario with sparse features, no dynamics, uneven-terrain. \\
			parking    & 3  & 16.8 & 1919.9  & 2810.6  & \textbf{\ding{72}\ding{73}\ding{73}\ding{73}\ding{73}} & Indoor, highly structured environment, rich features, few dynamics, flat-terrain. \\
			plaza      & 4  & 41.5 & 4329.1  & 6566.8  & \textbf{\ding{72}\ding{72}\ding{72}\ding{73}\ding{73}} & Indoor, multi-layer, lefts, high-dynamic crowds, flat-terrain. \\
			corridor   & 4  & 5.41 & 605.1   & 686.9   & \textbf{\ding{72}\ding{72}\ding{72}\ding{73}\ding{73}} & Indoor, highly repetitive geometric structures, no dynamics, flat-terrain. \\
			\midrule
			total      & 58 & 515.6 & 41792.6 & 51008.3 & \textbf{\ding{72}\ding{72}\ding{72}\ding{73}\ding{73}} & A total of 58 sequences covering common ground robot application scenarios and challenging elements. \\
			\bottomrule
	\end{tabular}}
	\label{table4}
\end{table*}

\begin{enumerate}
	\item[(i)] \textbf{open:} These sequences are collected along the shores of Bosten Lake in Xinjiang, China, and in the Urumqi gobi, which are uneven-terrain. Along the lakeshore, swaying reeds are abundant as unstable feature points, and LiDAR scans of the lake surface produce invalid reflections, which poses challenges for SLAM. The gobi is open and untraversed. Due to lack of distinctive features, Both LiDAR and visual SLAM often experience degradation.
	
	\item[(ii)] \textbf{rural:} It includes a large-scale, low-dynamic rural route and small-scale, beautiful flower beds strolling. The terrain is flat; however, the long travel distance often leads to accumulative errors.
	
	\item[(iii)] \textbf{urban:} The urban sequences are collected in Shenyang, China, including numerous highly-dynamic objects and various types of vehicles, which is a prominent research topic in SLAM. It comprises two sequences over 2 km, where many algorithms exhibit significant Z-axis drift.
	
	\item[(iv)] \textbf{aggressive:} We meticulously designed a series of routes incorporating aggressive motion, such as continuously walk up and down stairs, steep slopes with significant elevation changes, and rapid rotations (with a maximum rotational speed of 3.77 rad/s). Under these conditions, issues such as image blurring and LiDAR motion distortion frequently arise.
	
	\item[(v)] \textbf{campus:} The campus sequences are a highly structured scene with minimal dynamic objects and a flat-terrain; however, it includes water reflections after rain. This scene poses relatively low difficulty for most SLAM algorithms and is primarily designed to support researchers in evaluating system performance in campus environments, such as for inspection tasks.
	
	\item[(vi)] \textbf{park:} This is not a typical park walking route. Instead, it requires traversing grasslands and wooded areas within the park, featuring uneven-terrain, some of which also cover snow.
	
	\item[(vii)] \textbf{extreme:} The sequences include severe sandstorm and snow conditions (rarely found in most ground robot datasets), which enables the evaluation of algorithm performance under extreme weather conditions.
	
	\item[(viii)] \textbf{mixed:} Considering that many robotic systems in real-world applications must work between multiple operational scenarios, we design a series of multi-scenario mixed sequences, incorporating combinations such as indoor and outdoor environments, parks and urban areas, and parking and plazas. These sequences process the characteristics of multiple scenarios, presenting substantial challenges for SLAM algorithms.
	
	\item[(ix)] \textbf{smoke:} Artificial smoke is generated in the gobi scenario, and robot is deployed to navigate through it to evaluate SLAM algorithm performance under smoke occlusion. Additionally, the lack of distinguishable features further make trouble on SLAM algorithms.
	
	\item[(x)] \textbf{parking:} Three sequences are recorded in the underground parking to validate SLAM performance in this environment, where various loops are specifically designed to support research on indoor SLAM algorithms.
	
	\item[(xi)] \textbf{plaza:} Data for this category is collected in two large plazas, featuring highly-dynamic pedestrian, changing appearance, and ground reflections, which present substantial challenges for SLAM algorithms. Specifically, we record data across multiple floors and lift to evaluate relocation performance.
	
	\item[(xii)] \textbf{corridor:} This environment exists a highly-repetitive geometric structure, a typical degenerate scenario in LiDAR SLAM. Additionally, visual SLAM struggle in the long, dark corridors.
\end{enumerate}

\subsection{Ground truth}

\subsubsection{Localization ground truth:}
\begin{figure}[!th]
	\centering
	\includegraphics[width=0.5\textwidth]{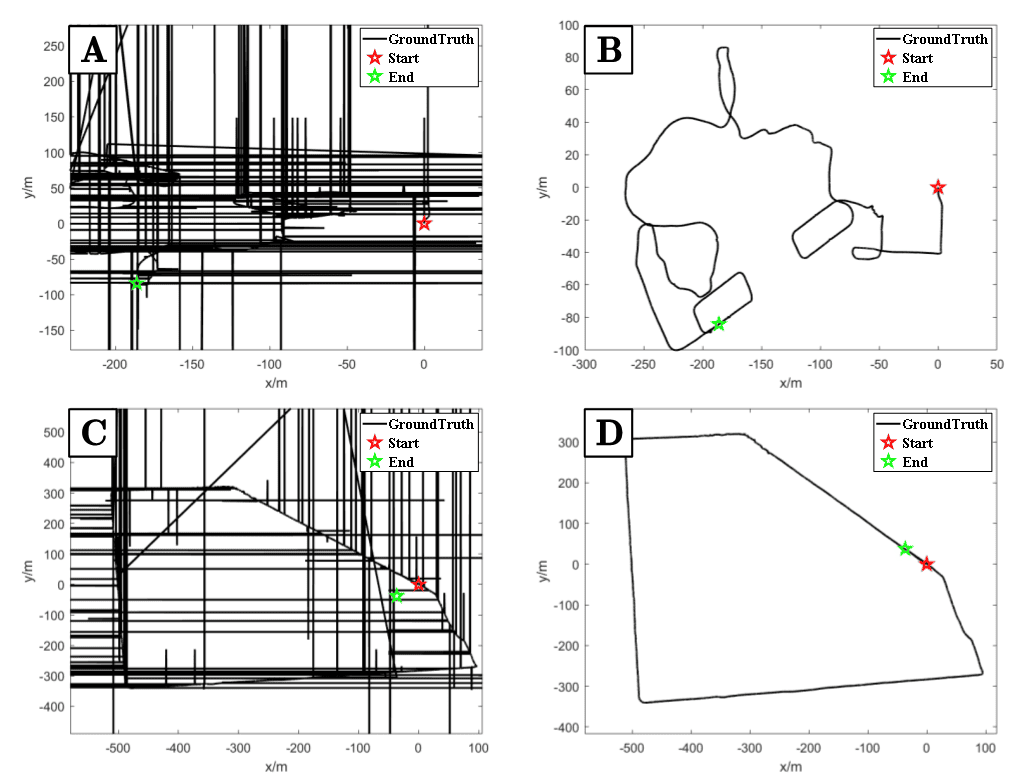}
	\caption{\textbf{Processed Outdoor Localization Ground Truths.} (A) and (C) show the raw RTK data for the aggressive\_05 and urban\_04 sequences, respectively, containing numerous outliers and significant errors; (B) and (D) show the localization ground truths refined using our method.}
	\label{figure6}
\end{figure}

\begin{figure}[!th]
	\centering
	\includegraphics[width=0.5\textwidth]{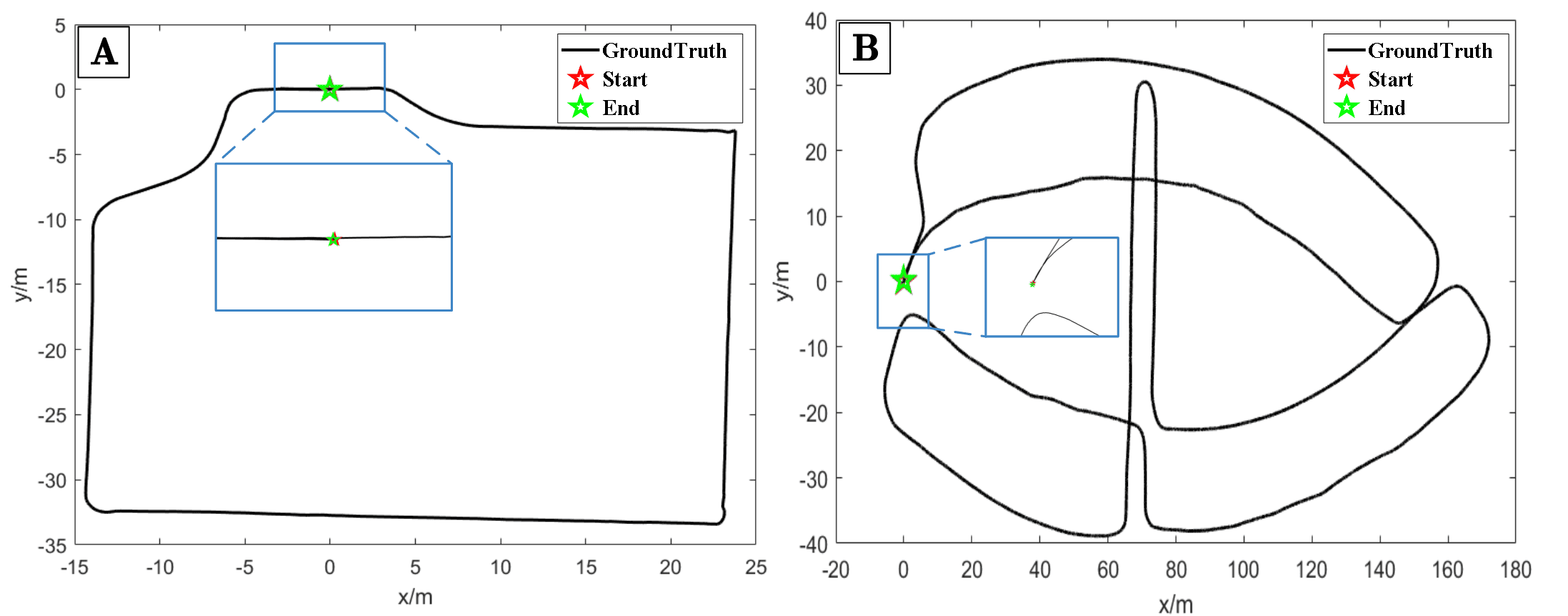}
	\caption{\textbf{Indoor Trajectory Ground Truth.} (A) Offline ground truth generated for the long corridor test sequence; (B) Offline ground truth generated for the parking test sequence. The zoom-in view highlights the high alignment between the start and end points, demonstrating the reliability of our method.}
	\label{figure7}
\end{figure}

For outdoor sequences, the original absolute poses of the robots are obtained from RTK. However, due to the notorious multipath and non-line-of-sight (NLOS) phenomenon (\cite{gnss1, gnss2}), the original data contain numerous outliers and significant errors, as shown in Figures~\ref{figure6} (A), (C). Therefore, the ground truth smooth kit in Section 4.3 is utilized to smooth the original data and generate high-quality localization ground truths, as shown in Figures~\ref{figure6} (B) and (D). For indoor sequences, our developed multi-sensor fusion SLAM algorithm (\cite{gr-loam}) is executed offline to generate accurate smooth trajectories ground truth (\cite{complex_urban, openLORIS, cid-sims}). Several sequences that returned to the start point are selected to verify the reliability of this ground truth generation method by analyzing the end-to-end errors. The results indicate that an end-to-end error of 0.031 m is observed after traversing 130 m in the long corridor test sequence, while an end-to-end error of 0.1 m is recorded after covering 601 m in the parking sequence. Therefore, we claim that this method is considered reliable. Figure~\ref{figure7} visually presents the ground truth generated by this method, along with the distances between the starting and ending points. For indoor-outdoor mixed sequences, our developed multi-sensor fusion SLAM is also executed offline, and when RTK data is available, RTK factor constraints are incorporated into the factor graph to enhance the reliability of the ground truth.

\subsubsection{Mapping ground truth:}
\begin{figure*}[!th]
	\centering
	\includegraphics[width=\textwidth]{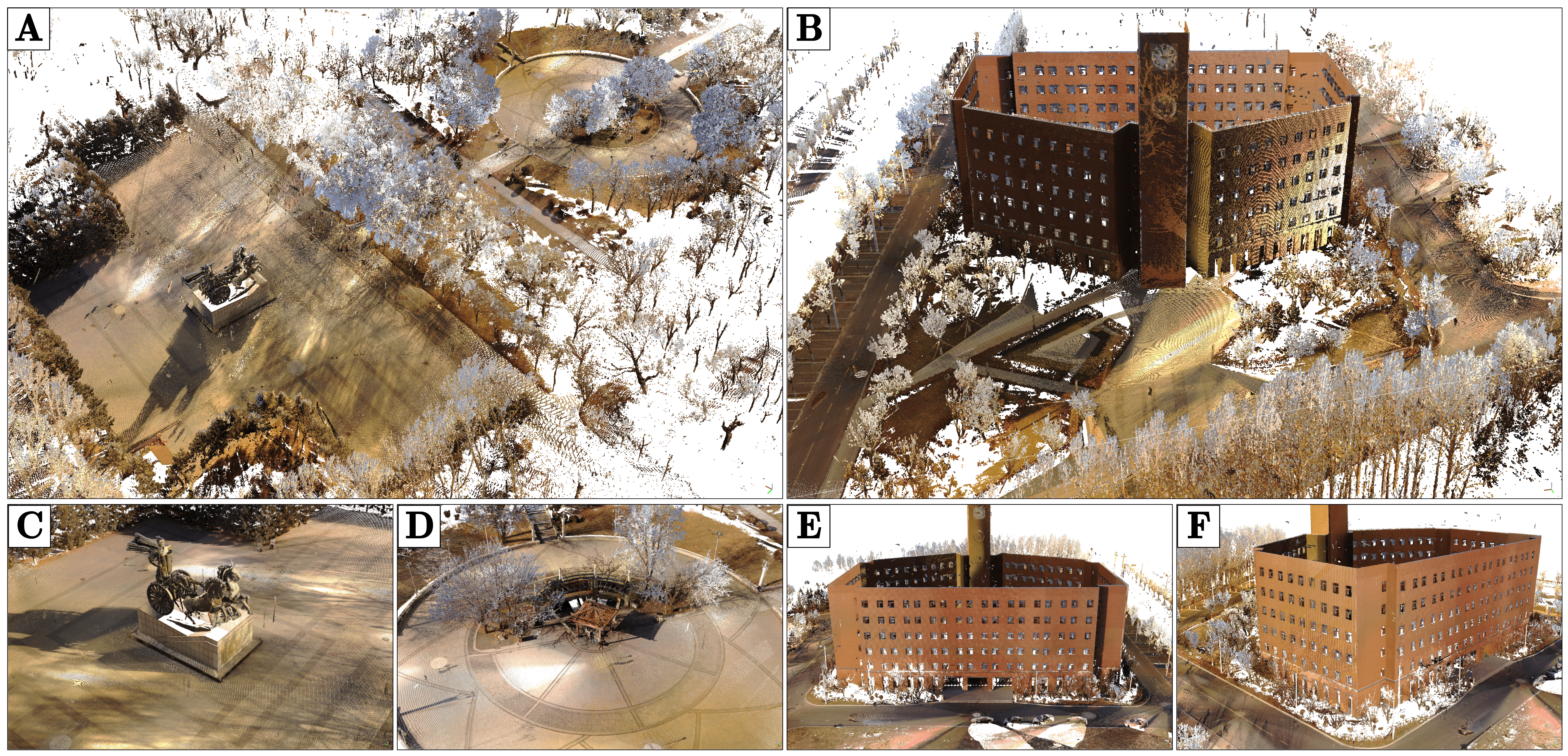}
	\caption{\textbf{Mapping Ground Truth.} (A) and (B) show the overall views of the park and dormitory building maps, respectively; (C)-(F) present zoomed-in partial views and snapshots from different perspectives.}
	\label{figure8}
\end{figure*}

To advance research and development in mapping algorithms, a FARO Focus Premium 70M 3D laser scanner\endnote{https://www.weiyang3d.com/en/h-col-124.html} is utilized to capture a 3D scene map of a park and a dormitory building, as illustrated in Figure~\ref{figure8}. This 3D laser scanner has a maximum visual range of 614 m and is capable of capturing 500,000 points per second, achieving a ranging error of $\pm$ 1 mm and a 3D positioning accuracy of 3.5 mm at a distance of 25 m. We match the scans offline and manually remove dynamic ghosting to construct accurate global map ground truths. Additionally, several handheld and ground robot sequences additionally provided for these scenarios, named as "mapping" in the open-source dataset, are specifically intended for map building and evaluation, whose routes are designed based on the scanning trajectories during ground truth construction.

\begin{figure}[!th]
	\centering
	\includegraphics[width=0.45\textwidth]{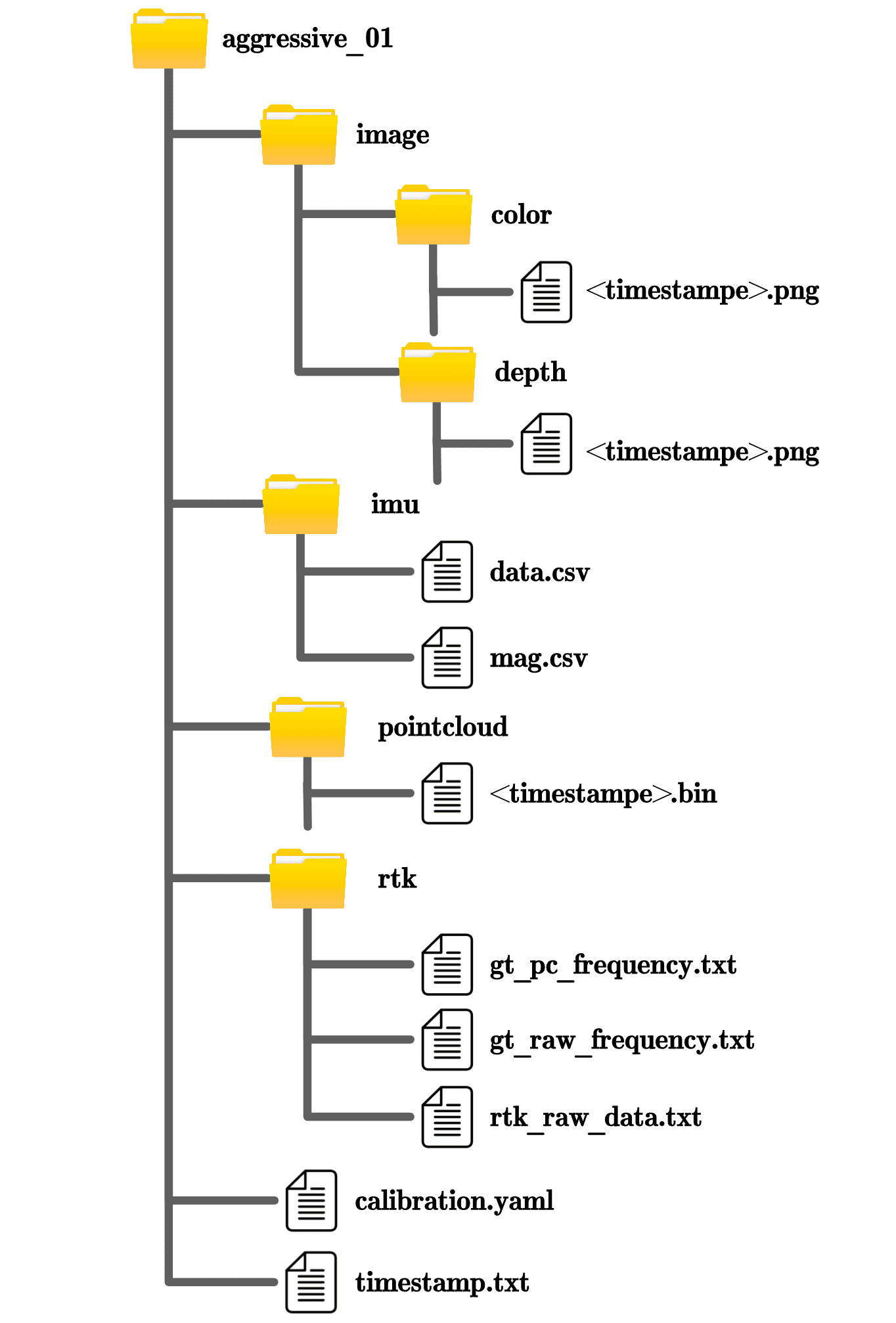}
	\caption{\textbf{The File Structure Converted by the \textit{bag2file.cpp} tool.}}
	\label{figure9}
\end{figure}

\subsubsection{Development kit:}
To enhance usability and facilitate evaluation, a series of development kits are also released on the M2UD website, which are implemented in C++ and built based on ROS. A brief overview of their main functions is provided as follows and more detailed usage instructions can be found on our website.
\begin{enumerate}
	\item[(i)] \textbf{Ground truth smooth:} The \textit{groundtruth\_create.cpp} file reads the RTK topic from the bag file and smooths the data. Eventually, it creates a \verb+GTCreate+ folder in the same directory and generates four files: timestamps, raw RTK data, smoothed ground truth values at the RTK frequency, and ground truth values at the LiDAR frequency. This process filters out outliers and noise in the RTK data, as illustrated in Figure~\ref{figure6}.
	
	\item[(ii)] \textbf{File format conversion:} The \textit{bag2file.cpp} file reads topics from the bags and converts color images, depth images, and point clouds into 8-bit \verb+.JPEG+ files and \verb+.bin+ binary files, respectively, naming them based on timestamps. It stores accelerometer, gyroscope, and magnetometer data from the IMU in a \verb+.csv+ file, the RTK data and smoothed trajectory ground truth in \verb+.txt+ files, and the sensor extrinsic and intrinsic parameters of the recording platform in a \verb+calibration.yaml+ file. The resulting file structure is illustrated in Figure~\ref{figure9}.
	
	\item[(iii)] \textbf{Time alignment:} The \textit{time\_align.cpp} file is primarily used to evaluate localization accuracy. Since some algorithms that adopt the keyframe strategy and incorporate a loop closure detection module (e.g., LIO-SAM \cite{lio-sam}) may output sparse trajectories for evaluation, the resulting trajectory may appear discontinuous when compared with the ground truth. This tool interpolates the algorithm's output based on the timestamps of the ground truth, enabling a more reasonable and continuous evaluation of localization accuracy.
	
	\item[(iv)] \textbf{RTK topic publisher:} The \textit{RTK\_process.cpp} file reads GGA or VTG protocol data from the serial port, then repackages and publishes the original position information, the number of GPS satellites, timestamps, signal quality, and other relevant data as RTK ROS topics.
	
	\item[(v)] \textbf{Localization evaluation:} To facilitate the evaluation of localization accuracy using the novel metric proposed in Section 5.1, we modify the widely used trajectory evaluation tool evo\endnote{https://github.com/MichaelGrupp/evo} in SLAM and introduce EA-evo. Users first need to export a frame-by-frame time consumption file in the format \verb+<timestamp time-consumption>+. Following the original style of evo, users only need to add the \verb+-ea time_consumption.txt+ parameter to apply efficiency-aware Drift rate and ATE for trajectory evaluation. Additionally, the \verb+-sf+ parameter can be used to specify the sensor frequency.
\end{enumerate}

\section{Data validation}

\subsection{Localization evaluation metrics}
We select three metrics to evaluate the localization performance of SLAM systems: the Root Mean Square Error (RMSE) of Absolute Trajectory Error (ATE), the Drift rate (Drift), and our proposed Efficiency-Aware Drift rate (EA-Drift).

\subsubsection{ATE:}
The RMSE of ATE is calculated by comparing the trajectory estimated by the SLAM algorithm with the ground truth trajectory provided in the dataset. Let $P_{1:n}, Q_{1:n}$ denote the time-aligned poses of the estimated trajectory and the ground truth trajectory, respectively. The RMSE of ATE between the two trajectories is defined as:
\begin{equation} \label{eq1}
	RMSE:=\sqrt{\frac{1}{n}\sum_{i=1}^n{\lVert P_{i}^{-1}SQ_i \rVert ^2}},
\end{equation}
where, $S$ represents the rigid body transformation obtained as the least squares solution that aligns the estimated trajectory $P_{1:n}$ to the ground truth trajectory $Q_{1:n}$, solved in closed form using the Horn method (\cite{ate}) and automatically aligned using evo. In our evaluation, we consider only the translation error, as the rotation error is represented as a translation error (\cite{tum_rgb-d}). ATE is used to evaluate the global consistency of SLAM trajectories.

\subsubsection{Drift:}
To evaluate the accumulative error relevant to the trajectory, we define the drift rate (Drift) as the ratio of the RMSE of ATE to the path length:
\begin{equation} \label{eq2}
	Drift:=\frac{\sqrt{\sum_{i=1}^n{\lVert P_{i}^{-1}SQ_i \rVert ^2}}}{L},
\end{equation}
where, $L$ is estimated path length.

\subsubsection{EA-Drift:}
We note that the ATE in~(\ref{eq1}) and the drift rate in~(\ref{eq2}) above consider only the localization accuracy of the SLAM algorithm, without accounting for its efficiency into the evaluation. However, real-time performance is an important requirement for the SLAM algorithm. Particularly, in practical robot applications, the SLAM algorithm, as part of the robot system, coexists with other functional algorithms, requiring that its time consumption is fully considered when selecting the SLAM algorithm deployed on mobile computers or even embedded systems. This is because an occasional frame loss may lead to robot collisions, or other safety hazards. Therefore, we propose a novel metric, Efficiency-Aware Drift rate (EA-Drift): 
\begin{equation} \label{eq3}
	EA-Drift:=\frac{\sqrt{\sum_{i=1}^n{e^{\frac{t_i-\frac{1}{f}}{f}}\lVert P_{i}^{-1}SQ_i \rVert ^2}}}{L},
\end{equation}
where, $t_i$ represents the time consumption of frame $i$ in ms, and $f$ indicates the sensor frequency. The evaluation metric combines the algorithm's time consumption and the accumulated localization error to form a comprehensive criterion, helping researchers choose a more suitable SLAM algorithm for practical applications. As described in Section 4.3, this evaluation has been integrated into evo for ease of use.

\subsection{Localization performance evaluation}
We meticulously select one sequence from each category for the SLAM localization evaluation, and the details of these 12 sequences are presented in Table~\ref{table5}. These sequences cover all difficulty levels and include the most common ground robot application scenarios to provide a comprehensive evaluation of SLAM algorithms. It should be noted that, except for the rural\_01 sequence that is no image data, while all other sequences contain LiDAR, camera, and IMU data. All algorithms fail after entering the lift in the plaza\_04 sequence; therefore, we just evaluate them before entering the lift.

\begin{table}[!t]
	\caption{12 Sequences for SLAM localization performance evaluation.}
	\centering
	\newcolumntype{C}[1]{>{\centering\arraybackslash}m{#1}}
	\newcolumntype{L}[1]{>{\raggedright\arraybackslash}m{#1}}
	\scalebox{0.8}{
		\begin{tabular}{C{2cm}C{2cm}C{2cm}C{2cm}}
			\toprule
			\textbf{Sequence} & \textbf{Duration (s)} & \textbf{Distance (m)} & \textbf{Difficulty} \\
			\midrule
			open\_08        & 2511.4 & 2789.9 & \ding{72}\ding{72}\ding{72}\ding{72}\ding{72} \\
			rural\_01       & 463.6  & 570.9  & \ding{72}\ding{73}\ding{73}\ding{73}\ding{73} \\
			urban\_05       & 2476.6 & 2807.9 & \ding{72}\ding{72}\ding{72}\ding{72}\ding{72} \\
			aggressive\_04  & 493    & 508.7  & \ding{72}\ding{72}\ding{72}\ding{73}\ding{73} \\
			campus\_05      & 563.6  & 667.6  & \ding{72}\ding{72}\ding{73}\ding{73}\ding{73} \\
			park\_02        & 569.6  & 602.8  & \ding{72}\ding{72}\ding{73}\ding{73}\ding{73} \\
			extreme\_03     & 412.9  & 442.7  & \ding{72}\ding{72}\ding{72}\ding{73}\ding{73} \\
			mixed\_02       & 554.4  & 641.9  & \ding{72}\ding{72}\ding{73}\ding{73}\ding{73} \\
			smoke\_02       & 223.6  & 240.8  & \ding{72}\ding{72}\ding{72}\ding{73}\ding{73} \\
			parking\_01     & 606.2  & 885.7  & \ding{72}\ding{73}\ding{73}\ding{73}\ding{73} \\
			plaza\_04       & 2215.9 & 3713.8 & \ding{72}\ding{72}\ding{72}\ding{73}\ding{73} \\
			corridor\_04    & 139.7  & 130.6  & \ding{72}\ding{72}\ding{72}\ding{73}\ding{73} \\
			\bottomrule
	\end{tabular}}
	\label{table5}
\end{table}

We test several state-of-the-art SLAM algorithms based on different principles to validate various data formats, including the feature-based LiDAR-only odometry A-LOAM (\cite{loam}), the direct-registration-based LiDAR-only odometry KISS-ICP (\cite{kiss-icp}), and the tightly-coupled LiDAR-inertial SLAM method LIO-SAM (\cite{lio-sam}) based on the graph optimization principle; the tightly-coupled LiDAR-inertial odometry FAST-LIO2 (\cite{fast-lio}), based on the filter principle; the sparse-direct-based visual odometry DSO (\cite{dso}); the feature-based visual-inertial SLAM method VINS-Mono \cite{vins-mono}; and the feature-based RGB-D SLAM method ORB-SLAM3 \cite{orb-slam3}.

\begin{table*}[!t]
	\caption{Benchmark results for 12 sequences in the M2UD dataset. The best results for each sequence are highlighted in \textbf{bold}. A dash ("-") indicates the failure of the algorithm’s execution. An "x" signifies a failure due to the RMSE over 50m or the Drift over 100\%.}
	\centering
	\newcolumntype{C}[1]{>{\centering\arraybackslash}m{#1}<{\centering}}
	\newcolumntype{L}[1]{>{\raggedright\arraybackslash}m{#1}}
	\scalebox{0.8}{
		\begin{tabular}{C{2cm}|C{2cm}|C{2cm}C{2cm}C{2cm}C{2cm}C{2cm}C{2cm}C{2cm}}
			\toprule
			\multicolumn{2}{c|}{\textbf{Method}} & \textbf{A-LOAM} & \textbf{KISS-ICP} & \textbf{LIO-SAM} & \textbf{FAST-LIO2} & \textbf{DSO} & \textbf{VINS-Mono} & \textbf{ORB-SLAM3} \\
			\midrule
			
			\textbf{Sequence} & \textbf{Metric} & LiDAR-only & LiDAR-only & LiDAR-inertial & LiDAR-inertial & Visual-only & Visual-inertial & RGB-D \\
			\midrule
			\multirow{3}{*}{open\_08} & APE & - & - & - & \textbf{2.24} & - & - & - \\
			& Drift & - & - & - & \textbf{3.59} & - & - & - \\
			& EA-Drift & - & - & - & \textbf{1.39} & - & - & - \\
			\midrule
			\multirow{3}{*}{rural\_01} & APE & 0.83 & 1.01 & 0.22 & \textbf{0.07} & - & - & - \\
			& Drift & 2.63 & 4.33 & 0.96 & \textbf{0.29} & - & - & - \\
			& EA-Drift & 16.07 & 1.87 & 0.49 & \textbf{0.11} & - & - & - \\
			\midrule
			\multirow{3}{*}{urban\_05} & APE & x & - & x & \textbf{31.89} & - & - & - \\
			& Drift & \textbf{41.62} & - & x & 51.27 & - & - & - \\
			& EA-Drift & x & - & x & \textbf{20.64} & - & - & - \\
			\midrule
			\multirow{3}{*}{aggressive\_04} & APE & 3.74 & 6.97 & 1.01 & \textbf{0.96} & 32.64 & 40.45 & 49.06 \\
			& Drift & 8.37 & 25.01 & 8.05 & \textbf{2.54} & x & - & x \\
			& EA-Drift & 59.29 & 12.58 & 4.10 & \textbf{1.01} & 89.73 & 89.73 & 58.07 \\
			\midrule
			\multirow{3}{*}{campus\_05} & APE & \textbf{2.22} & 2.95 & 4.66 & 4.93 & 32.28 & x & 9.32 \\
			& Drift & \textbf{4.93} & 10.00 & 34.54 & 16.44 & 43.65 & x & 29.17 \\
			& EA-Drift & 19.08 & \textbf{4.27} & 16.51 & 6.58 & 18.05 & x & 14.60 \\
			\midrule
			\multirow{3}{*}{park\_02} & APE & 5.90 & - & 4.57 & \textbf{4.16} & 38.82 & - & - \\
			& Drift & 12.44 & - & 35.53 & \textbf{14.75} & x & - & - \\
			& EA-Drift & x & - & 22.65 & \textbf{6.10} & 59.43 & - & - \\
			\midrule
			\multirow{3}{*}{extreme\_03} & APE & 1.16 & 1.58 & 0.28 & \textbf{0.14} & 6.58 & 22.57 & 10.78 \\
			& Drift & 11.11 & 20.78 & 4.06 & \textbf{2.08} & 22.73 & 99.84 & 94.66 \\
			& EA-Drift & 81.53 & 9.20 & 2.08 & \textbf{0.82} & 31.76 & 57.70 & 45.92 \\
			\midrule
			\multirow{3}{*}{mixed\_02} & APE & 4.24 & 3.12 & 1.85 & \textbf{1.34} & 23.80 & - & - \\
			& Drift & 8.17 & 9.27 & 12.13 & \textbf{3.31} & 51.92 & - & - \\
			& EA-Drift & 87.60 & 4.47 & 6.07 & \textbf{1.34} & 21.35 & - & - \\
			\midrule
			\multirow{3}{*}{smoke\_02} & APE & 5.56 & 0.94 & 0.91 & \textbf{0.73} & - & 32.01 & 24.68 \\
			& Drift & 27.25 & 5.10 & 11.43 & \textbf{4.01} & - & x & 92.53 \\
			& EA-Drift & 40.86 & 2.05 & 5.06 & \textbf{1.55} & - & x & 49.87 \\
			\midrule
			\multirow{3}{*}{parking\_01} & APE & 0.65 & 0.37 & 0.55 & \textbf{0.21} & 18.56 & 31.31 & 22.75 \\
			& Drift & 3.43 & 5.97 & 4.84 & \textbf{1.89} & x & x & x \\
			& EA-Drift & 28.70 & 2.38 & 2.46 & \textbf{0.75} & 42.26 & x & 57.99 \\
			\midrule
			\multirow{3}{*}{plaza\_04} & APE & \textbf{5.33} & 7.17 & 7.55 & 20.03 & - & - & - \\
			& Drift & \textbf{11.78} & 33.38 & 36.69 & x & - & - & - \\
			& EA-Drift & x & \textbf{13.81} & 57.88 & 60.36 & - & - & - \\
			\midrule
			\multirow{3}{*}{corridor\_04} & APE & \textbf{1.00} & - & - & - & - & 12.03 & 2.90 \\
			& Drift & \textbf{16.04} & - & - & - & - & x & 42.70 \\
			& EA-Drift & 22.01 & - & - & - & - & 75.76 & \textbf{19.97} \\
			\bottomrule
	\end{tabular}}
	\label{table6}
\end{table*}
\begin{figure*}[!th]
	\centering
	\includegraphics[width=\textwidth]{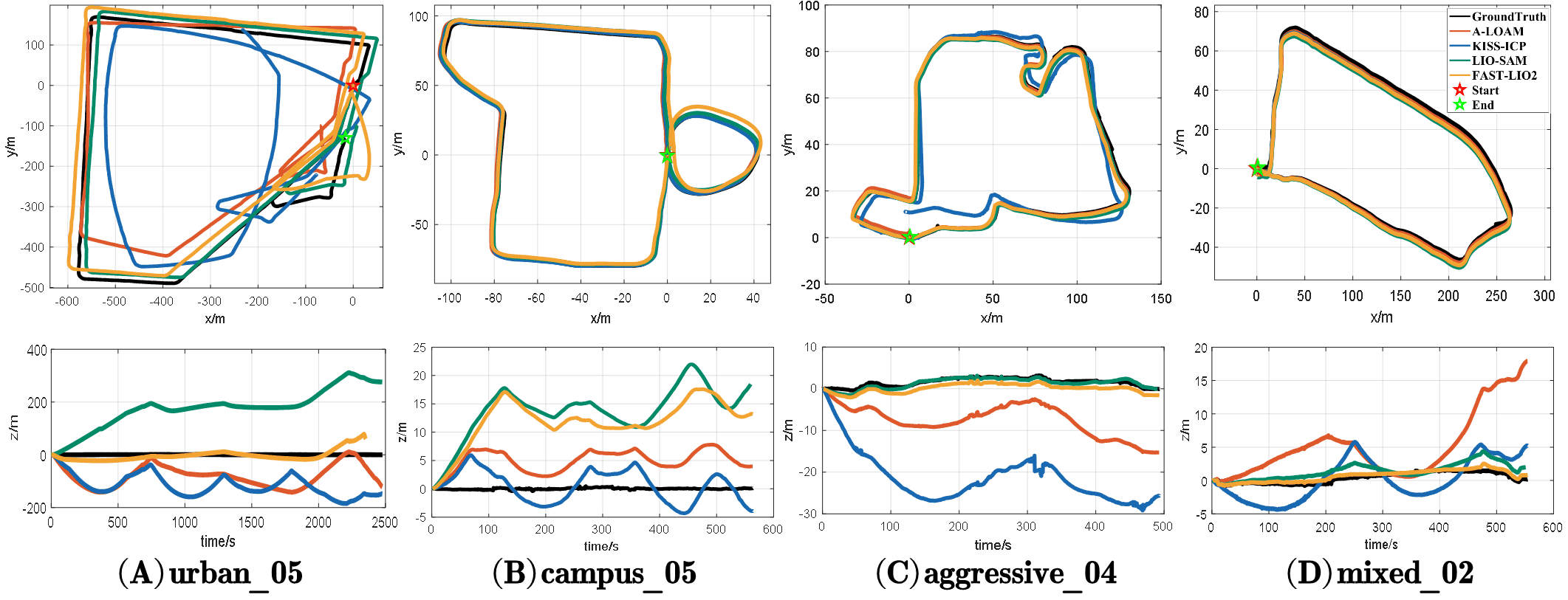}
	\caption{\textbf{Representative Sequence Trajectories.} The trajectory results indicate that the LiDAR SLAM algorithm frequently exhibits significant Z-axis drift on our dataset, leading to dissatisfactory localization accuracy.}
	\label{figure10}
\end{figure*}

\begin{figure*}[!th]
	\centering
	\includegraphics[width=\textwidth]{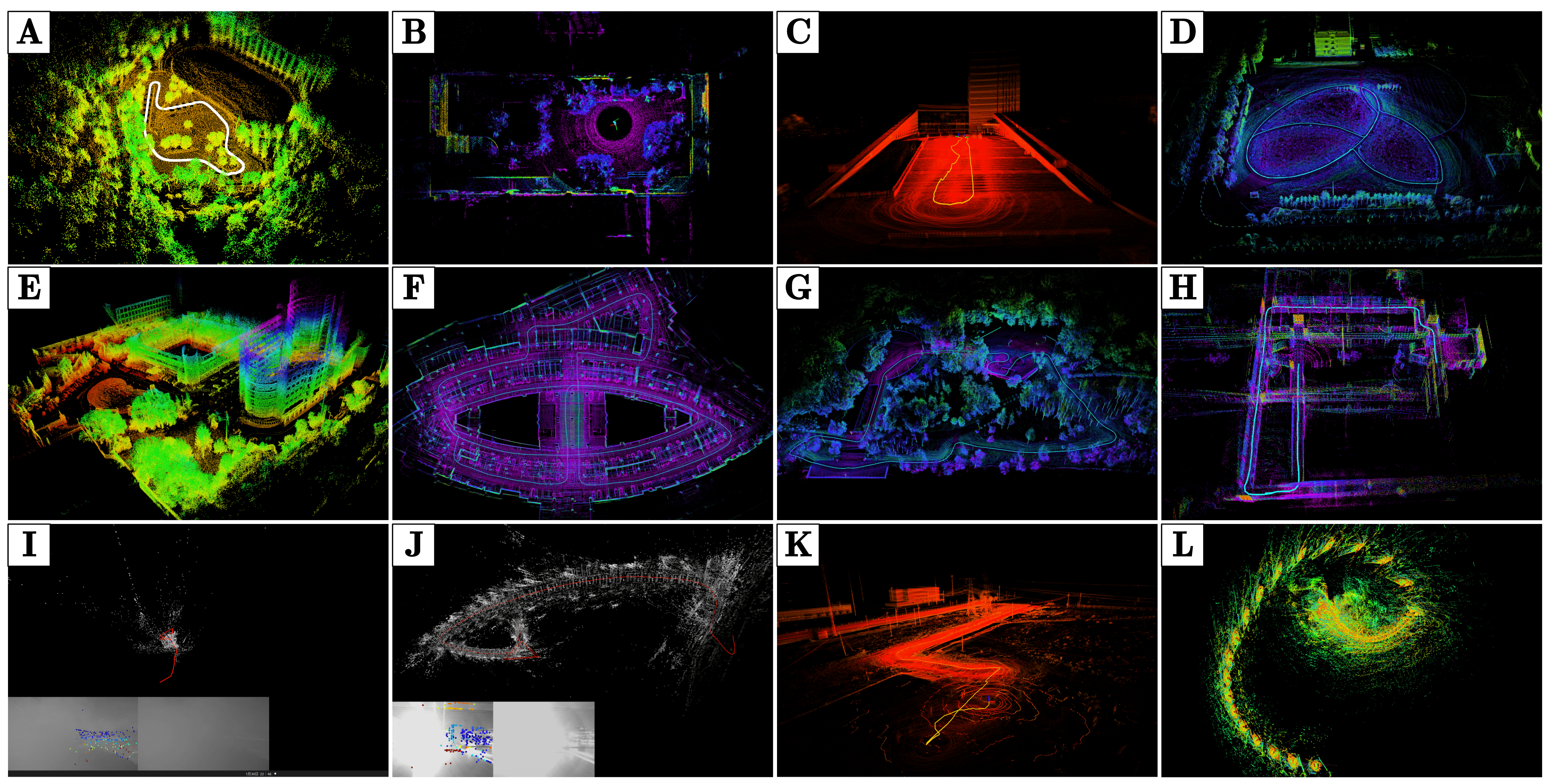}
	\caption{\textbf{Additional Test Results of SLAM Algorithms on the M2UD.} (A)–(G) present the maps successfully constructed by the algorithms. (H)–(L) highlight failure cases.}
	\label{figure11}
\end{figure*}

The benchmark results for 12 sequences are presented in Table~\ref{table6}. In the outdoor and large-scale sequences, such as open\_08 and urban\_05, the visual SLAM algorithm is often affected by adverse
factors such as overexposure, occlusion, and dynamic objects, leading to a system crash. In addition, since visual-based methods typically require favourable initialization, poor initialization may result in large errors. The initialization of DSO is relatively simple, but due to the lack of loop detection and global optimization, it exhibits large accumulative errors. ORB-SLAM3 uses depth information and includes loop detection and relocation modules, yielding relatively good results. However, in general, the LiDAR SLAM method is more accurate and robust than the visual SLAM method in most cases. A-LOAM successfully complete most sequences, although with slightly lower accuracy. However, in the open\_08 sequence, A-LOAM thoroughly failed due to the scene being too sparse and lacking effective features for registration. Additionally, as the system runs, the execution time of A-LOAM increases, leading to frame loss and a higher EA-Drift, which suggests that A-LOAM may pose security risks in practical applications. KISS-ICP directly uses raw point clouds for registration and can handle most sequences, but is generally less robust in highly dynamic, large-scale, unstructured environments. LIO-SAM and FAST-LIO2 tightly couple LiDAR and IMU to achieve high localization accuracy. Their EA-Drift performance is also excellent, demonstrating their advantages in both accuracy and efficiency, making them suitable alternatives to SLAM algorithms in practical applications. When returning to a historical scenes, LIO-SAM can activate the loop detection module to correct trajectory deviations, which contributes to improve global consistency. However, it may thoroughly fail in degraded environments such as open areas and long corridors. FAST-LIO2 achieves optimal results in most sequences. However, our dataset still presents significant challenges to it. For example, due to the influence of highly dynamic objects in urban\_05, it exhibits a large accumulative error and fails in the degenerate corridor\_04 sequence, which highlights the challenges that our dataset poses to existing SLAM algorithms.

In addition, although LiDAR SLAM is generally more accurate than visual methods, it often exhibits significant Z-axis drift due to the sparse nature of LiDAR data. Figure~\ref{figure10} presents trajectories across several representative sequences. The results indicate that even the trajectories estimated by these methods matches the ground truth in the XY direction, its Z-axis exhibits a significant drift, leading to a decline in localization accuracy, highlighting the challenges our dataset poses to existing LiDAR SLAM algorithms.

Figure~\ref{figure11} presents the test results of additional algorithms on the M2UD, including LeGO-LOAM (\cite{lego-loam}), DLIO (\cite{dlio}), Point-LIO (\cite{point-lio}), GR-LOAM (\cite{gr-loam}), FAST-LIVO (\cite{fast-livo}), DLO (\cite{dlo}), CT-ICP (\cite{ct-icp}), SVO (\cite{svo}), and RTAB-MAP (\cite{trab-map}). Figure~\ref{figure11} (A)–(G) present the maps successfully constructed by the algorithms. Figure~\ref{figure11} (H)–(L) highlight cases where the algorithms fail. With its high-bandwidth design, Point-LIO achieves accurate map construction on the aggressive\_07 rapid-rotation sequence, but the map still exhibits visible stratification as shown in Figure~\ref{figure11} (B). Figure~\ref{figure11} (H) visually illustrates the degradation of FAST-LIO2 on the corridor\_04 sequence. Figure~\ref{figure11} (I) and (J) illustrate the impact of smoke or overexposure on visual SLAM algorithms. Methods based on direct registration, such as DLIO and DLO, often fail in open environments as shown in Figure~\ref{figure11} (K), as observed in the LIO-SAM run shown in Figure~\ref{figure11} (L). Additional algorithm results and videos are available on the dataset website: \href{https://yaepiii.github.io/M2UD/}{https://yaepiii.github.io/M2UD/}.

\begin{table*}[!t]
	\caption{Test results for two mapping sequences in the M2UD. The best mapping results for each sequence are highlighted in \textbf{bold}. A dash ("-") indicates that the algorithm failed to execute.}
	\centering
	\newcolumntype{C}[1]{>{\centering\arraybackslash}m{#1}<{\centering}}
	\newcolumntype{L}[1]{>{\raggedright\arraybackslash}m{#1}}
	\scalebox{1.0}{
		\begin{tabular}{C{2cm}|C{2cm}|C{2cm}C{2cm}C{2cm}C{2cm}C{2cm}}
			\toprule
			\multicolumn{2}{c|}{\textbf{Method}} & \textbf{LIO-SAM} & \textbf{FAST-LIO2} & \textbf{Point-LIO} & \textbf{SLAMesh} & \textbf{ImMesh} \\
			\midrule
			
			\textbf{Sequence} & \textbf{Metric} & Point cloud & Point cloud & Point cloud & Mesh & Mesh \\
			\midrule
			\multirow{4}{*}{\textbf{mapping\_01}} & AC$\downarrow$ & \textbf{0.19} & 0.23 & 0.21 & 0.21 & 0.22 \\
			& CD$\downarrow$ & 0.20 & 0.68 & \textbf{0.11} & 0.22 & 0.18 \\
			& AWD$\downarrow$ & 0.74 & \textbf{0.49} & 1.05 & 0.73 & 0.85 \\
			& SCS$\downarrow$ & 0.62 & 0.54 & \textbf{0.48} & 0.56 & 0.49 \\
			\midrule
			\multirow{4}{*}{\textbf{mapping\_02}} & AC$\downarrow$ & \textbf{0.20} & 0.21 & 0.21 & 0.21 & - \\
			& CD$\downarrow$ & 0.13 & \textbf{0.06} & 0.22 & 0.16 & - \\
			& AWD$\downarrow$ & 1.30 & 1.49 & \textbf{1.18} & 1.21 & - \\
			& SCS$\downarrow$ & 0.51 & 0.41 & 0.56 & \textbf{0.26} & - \\
			\bottomrule
	\end{tabular}}
	\label{table7}
\end{table*}

\begin{figure*}[!th]
	\centering
	\includegraphics[width=\textwidth]{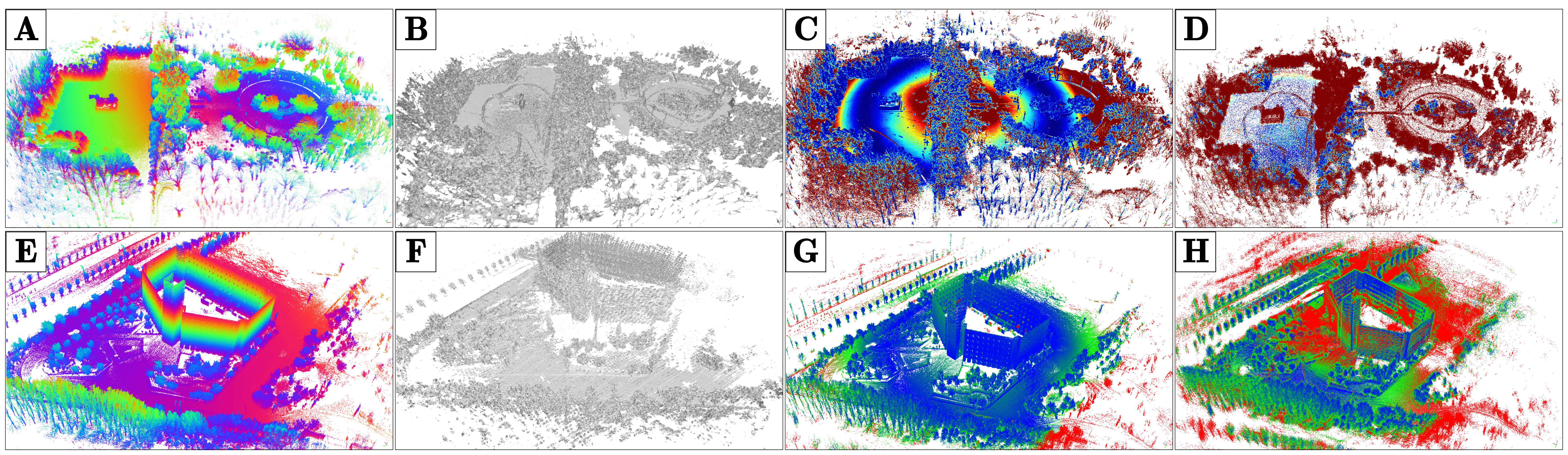}
	\caption{\textbf{Mapping Evaluation Results Visualizations.} (A) and (E) show the built point cloud maps; (B) and (F) display the mesh maps created by ImMesh and SLAMesh for mapping\_01 and mapping\_02, respectively. (C) and (D) present the evaluation results visualized using MapEval; (G) and (H) show the evaluation results visualized with CloudCompare\protect\endnote{https://www.cloudcompare.org/}.}
	\label{figure12}
\end{figure*}

\subsection{Mapping evaluation metrics}
We utilize the advanced map evaluation tool (\cite{mapeval}), and select four evaluation metrics to assess mapping performance:
\subsubsection{Accuracy (AC):}
Measures the Euclidean error of correctly reconstructed corresponding points within a predefined threshold:
\begin{equation} \label{eq4}
	AC:=\frac{1}{\left| \mathcal{C}_{\tau} \right|}\sum_{\left( p_{i}^{g},q_{j}^{e} \right) \in \mathcal{C}_{\tau}}{f}\left( \lVert p_{i}^{g}-q_{j}^{e} \rVert <\tau \right),
\end{equation}
where, $p_{i}^{g}$ denotes the $i$-th point in the ground truth map $\mathcal{M}_g$; $q_{j}^{e}$ denotes the $j$-th point in the estimated map $\mathcal{M}_e$; $f(\cdot)$ is the indicator function; $\mathcal{C}_{\tau}$ represents the number of corresponding points within the threshold. we set the threshold $\tau =0.1$.

\subsubsection{Chamfer Distance (CD):}
Provides a symmetric metric for the average nearest-point distance between two point clouds:
\begin{equation} \label{eq5}
	\begin{split}
	CD\left( \mathcal{M}_g,\mathcal{M}_e \right) &=\frac{1}{N_g}\sum_{p_{i}^{g}\in \mathcal{M}_g}{\underset{p_{j}^{e}\in \mathcal{M}_e}{\min}\lVert p_{i}^{g}-p_{j}^{e} \rVert} \\
	&+\frac{1}{N_e}\sum_{p_{j}^{e}\in \mathcal{M}_e}{\underset{p_{i}^{g}\in \mathcal{M}_g}{\min}\lVert p_{j}^{e}-p_{i}^{g} \rVert}.
	\end{split}
\end{equation}

CD considers the bidirectional Euclidean distance between all points in the the ground truth $\mathcal{M}_g$ and estimated map $\mathcal{M}_e$, which enables it to capture local details more effectively than AC (\cite{dac}). However, this method is sensitive to outliers and exhibits high computational complexity, particularly for large-scale point clouds. Therefore, we employ the two most recent metrics proposed in \cite{mapeval} to evaluate mapping performance.

\subsubsection{Average Wasserstein Distance (AWD):}
First, the ground truth and the estimated map are represented as Gaussian Mixture Model (GMM), with the point cloud divided into same voxel sets $V=\left\{ v_1,v_2,...,v_M \right\} $. The voxel size is set to 3 m. In each voxel $v$, a Gaussian distribution is used to approximate the distribution of points $P_i$, characterized by their mean $\mu_i$ and covariance $\sigma_i$.
\begin{equation} \label{eq6}
	\mu _i=\frac{1}{\left| P_i \right|}\sum_{p\in P_i}{p},\ \Sigma _i=\frac{1}{\left| P_i \right|-1}\sum_{p\in P_i}{\left( p-\mu _i \right) \left( p-\mu _i \right) ^T}.
\end{equation}

Next, the $\mathcal{L}2$ Wasserstein distance is calculated between the corresponding voxel distributions:
\begin{equation} \label{eq7}
	\begin{split}
	&W^2( \mathcal{N}_{i}^{g},\mathcal{N}_{i}^{e} ) \\
	&=\lVert \mu _{i}^{g}-\mu _{i}^{e} \rVert ^2+tr( \Sigma _{i}^{g}+\Sigma _{i}^{e}-2( {\Sigma _{i}^{e}}^{\frac{1}{2}}\Sigma _{i}^{g}{\Sigma _{i}^{e}}^{\frac{1}{2}} ) ^{\frac{1}{2}}),
	\end{split}
\end{equation}
where, $\mathcal{N}_{i}^{g}$ and $\mathcal{N}_{i}^{e}$ represent the Gaussian distributions of the $i$-th voxelin the ground truth map and the estimated map, respectively; $tr(\cdot)$ denotes the trace of the matrix. Finally, the AWD for all $M$ voxels is defined as: $AWD:=\frac{1}{M}\sum\limits_{i=1}^M{W\left( \mathcal{N}_{i}^{g},\mathcal{N}_{i}^{e} \right)}$. AWD provides a global measure of map accuracy, capturing both the displacement between point distributions (via the mean) and the differences in local structure (via covariance). The metric is robust to noise and variations in point density.

\subsubsection{Spatial Consistency Score (SCS):}
To evaluate the local consistency of maps, we use the SCS: $SCS:=\frac{1}{M}\sum\limits_{i=1}^M{\frac{\sigma \left( W_{N\left( i \right)} \right)}{\mu \left( W_{N\left( i \right)} \right)}}$, where $W_{N\left( i \right)}$ is the set of Wasserstein distance of the neighboring voxels of $v_i$; $\sigma(\cdot)$ and $\mu(\cdot)$ represent the standard deviation and mean. The lower SCS, the more consistent the mapping error between neighboring regions, reflecting better local consistency.

\subsection{Mapping performance evaluation}
We select three SLAM algorithms based on point cloud map representation: LIO-SAM (\cite{lio-sam}), FAST-LIO2 (\cite{fast-lio}), and Point-LIO \cite{point-lio}, and two SLAM systems based on mesh map representation: SLAMesh (\cite{slamesh}) and ImMesh (\cite{immesh}) to perform mapping performance evaluation on two mapping sequences of M2UD. For the meshing methods, we first sample 1,000,000 points uniformly within the mesh and then compare them with the ground truth.

The quantitative and qualitative results are presented in Table~\ref{table7} and Figure~\ref{figure12}, respectively. The three LiDAR-inertial SLAM algorithms all achieve better results in terms of mapping. Among them, FAST-LIO2 and Point-LIO perform better in terms of CD, AWD, and SCS, reflecting their superior mapping detail representation and localization accuracy. Although LIO-SAM may neglect some details due to its keyframe strategy, resulting in suboptimal results, its ability of loop correction enables for a more globally consistent map, achieving a higher AC. As shown in Figure~\ref{figure12} (B) and (F), sparse-channel LiDAR presents a significant challenge to mesh-based methods, leading to more holes and affecting the quality of map construction. Additionally, ImMesh fails on mapping\_02 due to odometry drift, demonstrating that our mapping sequence requires both localization and mapping performance, which motivates further research on advanced localization and mapping methods in the future.

\section{Conclusion}
This paper presents a multi-modal, multi-scenario, uneven-terrain SLAM dataset for ground robots, providing several common application scenarios captured by 3D LiDAR, RGB-D cameras, IMU, and GNSS, along with accurate localization and mapping ground truth. The dataset contains 58 sequences across 12 categories, with a total length of more than 50 km, including extensive degraded situations, aggressive motion, and uneven terrain, filling the gap left by most existing ground robot datasets that only contain planar motion, which presents an open challenge to existing SLAM algorithms. We also open-source a set of development kits, including ground truth smooth, data format conversion, and time alignment, to facilitate research and further development of the dataset. We conduct extensive experiments with state-of-the-art methods and evaluated LiDAR, visual, and multi-sensor fusion SLAM algorithms using traditional and the proposed novel localization metrics EA-Drift, as well as mapping performance evaluation using our map ground truth to demonstrate the usability of our dataset. We believe our dataset will help advance the development, benchmarking, and practical application of SLAM algorithms for ground robots. In the future, we plan to assign 3D semantic labels to point clouds and images for use in navigable area and terrain analysis. Furthermore, we aspire to develop robust SLAM algorithms capable of overcoming these challenges in future research.

\begin{acks}
	This work was supported by the National Natural Science Foundation of China (62403458, U20A20201), National Key R\&D Program of China (2024YFB4709000), and Shenyang Science and Technology Program(23-407-3-38).
\end{acks}

\begin{dci}
The author(s) declared no potential conflicts of interest with respect to the research, authorship, and/or publication of this article.
\end{dci}

\begin{funding}
The author(s) disclosed receipt of the following financial support for the research, authorship, and/or publication of this article: This work was supported by the National Natural Science Foundation of China (62403458, U20A20201), National Key R\&D Program of China (2024YFB4709000), and Shenyang Science and Technology Program(23-407-3-38).
\end{funding}

\section*{ORCID IDs}

Yanpeng Jia ~\orcidlink{0009-0008-1295-7439} \href{https://orcid.org/0009-0008-1295-7439}{https://orcid.org/0009-0008-1295-7439} \\
Shiliang Shao ~\orcidlink{0000-0002-4512-167X} \href{https://orcid.org/0000-0002-4512-167X}{https://orcid.org/0000-0002-4512-167X} \\
Ting Wang ~\orcidlink{0000-0003-2616-3150} \href{https://orcid.org/0000-0003-2616-3150}{https://orcid.org/0000-0003-2616-3150}

\begin{sm}
Supplemental material for this article is available online.
\end{sm}

\theendnotes

%Harvard (name/date)
\bibliographystyle{SageH}
%Vancouver (numbered)
\bibliographystyle{SageV}
\bibliography{ref.bib}
\balance

\end{document}